\documentclass{article}

\usepackage[main, final]{neurips_2025}

\usepackage[utf8]{inputenc}
\usepackage[T1]{fontenc}
\usepackage{hyperref}
\usepackage{url}
\usepackage{booktabs}
\usepackage{amsfonts}
\usepackage{nicefrac}
\usepackage{microtype}
\usepackage{xcolor}
\usepackage{amsmath,amssymb,graphicx,color,algorithm,subfig, algpseudocode,booktabs,bm,relsize,enumitem,multirow,amsthm,epsfig,caption}
\usepackage[export]{adjustbox}
\usepackage{lscape}
\usepackage{multirow}
\usepackage{graphicx}
\usepackage{booktabs}
\usepackage{makecell}

\usepackage{microtype}
\usepackage{graphicx}
\usepackage{hyperref}

\usepackage{amsmath}
\usepackage{amssymb}
\usepackage{mathtools}
\usepackage{amsthm}
\usepackage{wrapfig}

\usepackage{booktabs,siunitx,makecell,array}
\usepackage[capitalise,noabbrev,nameinlink]{cleveref}
\creflabelformat{equation}{#2\textup{#1}#3}
\creflabelformat{section}{#2\textup{#1}#3}
\creflabelformat{appendix}{#2\textup{#1}#3}
\creflabelformat{figure}{#2\textup{#1}#3}
\crefname{algocf}{Algorithm}{Algorithms}
\Crefname{algocf}{Algorithm}{Algorithms}

\definecolor{DarkBlue}{rgb}{0,0.08,0.45}

\hypersetup{
colorlinks=true,
linkcolor=DarkBlue,
citecolor=DarkBlue,
filecolor=DarkBlue,
urlcolor=DarkBlue}

\usepackage{amsmath,amssymb,graphicx,color,algorithm,subfig, algpseudocode,booktabs,bm,relsize,enumitem,multirow,amsthm,epsfig,caption}
\usepackage[export]{adjustbox}
\usepackage{lscape}
\usepackage{multirow}
\usepackage{graphicx}
\usepackage{booktabs}
\usepackage{makecell}

\newsavebox{\measurebox}

\title{Coarse-to-fine Q-Network with Action Sequence \\ for Data-Efficient Reinforcement Learning}

\author{
  Younggyo Seo\\
  UC Berkeley \\
  \texttt{mail@younggyo.me} \\
  \And
  Pieter Abbeel \\
  UC Berkeley \\
  \texttt{pabbeel@cs.berkeley.edu}
}

\begin{document}

\maketitle

\begin{abstract}
  Predicting a sequence of actions has been crucial in the success of recent behavior cloning algorithms in robotics.
  Can similar ideas improve reinforcement learning (RL)?
  We answer affirmatively by observing that incorporating action sequences when predicting ground-truth return-to-go leads to lower validation loss.
  Motivated by this, we introduce Coarse-to-fine Q-Network with Action Sequence (CQN-\textbf{AS}), a novel value-based RL algorithm that learns a critic network that outputs Q-values over a sequence of actions, i.e., explicitly training the value function to learn the consequence of executing action sequences.
  Our experiments show that CQN-\textbf{AS} outperforms several baselines on a variety of sparse-reward humanoid control and tabletop manipulation tasks from BiGym and RLBench. Code is available at: \url{https://younggyo.me/cqn-as/}
\end{abstract}

\begin{figure*}[h]
\centering
\vspace{-0.15in}
\includegraphics[width=0.32\linewidth]{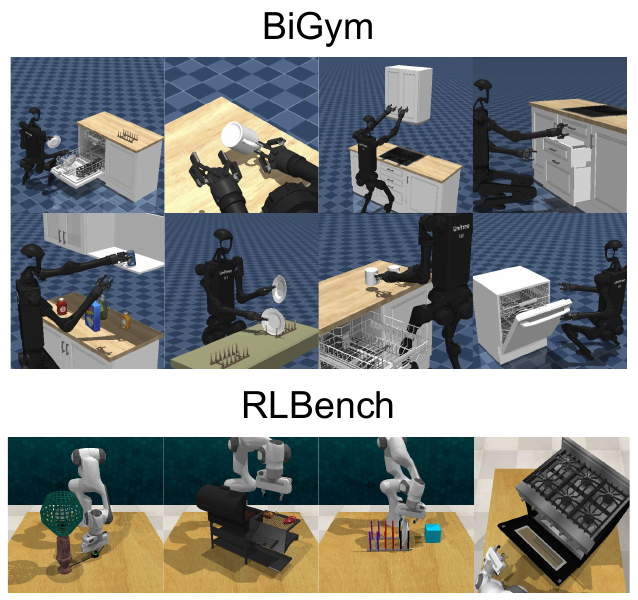}
\includegraphics[width=0.64\linewidth]{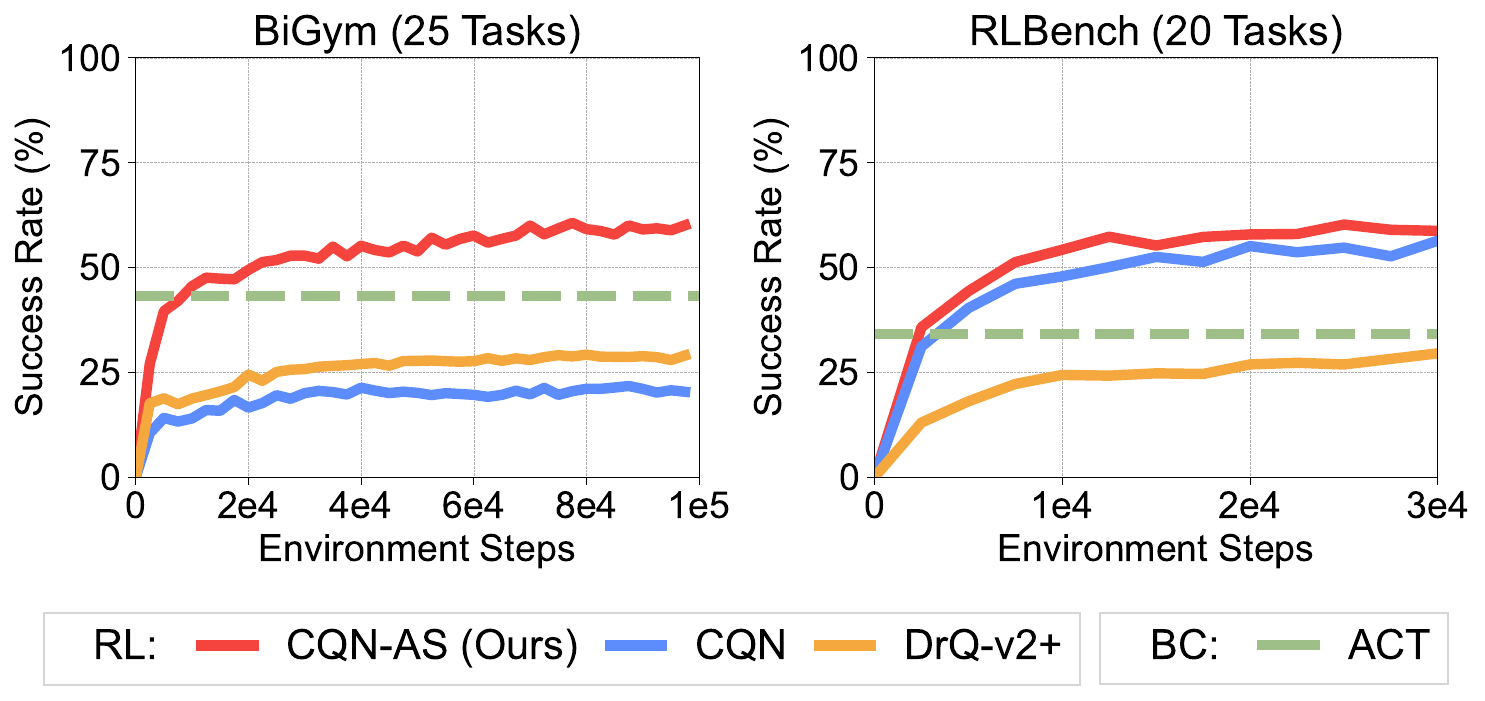}
\caption{\textbf{Summary of results.} Coarse-to-fine Q-Network with \textbf{A}ction \textbf{S}equence (CQN-\textbf{AS}) learns a critic network with action sequence.
CQN-\textbf{AS} outperforms various RL and BC baselines such as CQN \citep{seo2024continuous}, DrQ-v2+ \citep{yarats2022mastering}, and ACT \citep{zhao2023learning} on 45 robotic tasks from BiGym \citep{chernyadev2024bigym} and RLBench \citep{james2020rlbench}.}
\label{fig:summary_of_results}
\end{figure*}

\section{Introduction}
Predicting action sequences from expert trajectories is a key idea in recent successful behavior cloning (BC; \citealt{pomerleau1988alvinn}) approaches in robotics. This has enabled policies to effectively approximate the noisy, multi-modal distribution of expert demonstrations \citep{zhao2023learning,chi2023diffusion}.
Can this idea similarly be useful for reinforcement learning (RL)?

Our initial finding is affirmative: we make an intriguing observation that using action sequences can enhance value learning.
Specifically, with humanoid demonstrations from BiGym \citep{chernyadev2024bigym}, we train regression models that predict the ground-truth return-to-go, i.e., the sum of discounted future rewards from the timestep $t$, given the current observation and action.
In \cref{fig:return_to_go_prediction}, we find that using an action sequence $\mathbf{a}_{t:t+K} = \{\mathbf{a}_{t}, ..., \mathbf{a}_{t+K-1}\}$ as input results in lower validation losses than using a single-step action $\mathbf{a}_{t}$.
We hypothesize this is because action sequences, which can correspond to behavioral primitives such as \textit{going straight}, make it easier for the model to learn the long-term outcomes compared to evaluating the effect of individual single-step actions (see \cref{appendix:motivating_experiments} for additional analysis based on a 2D Point-mass environment).

Building on this observation, we train actor-critic algorithms \citep{haarnoja2018soft,fujimoto2018addressing} with action sequence on \texttt{stand} task from HumanoidBench \citep{sferrazza2024humanoidbench}.
Specifically, we train the actor to output action sequence and the critic to take action sequence as inputs instead of single-step actions.
However, we find these algorithms with action sequences suffer from severe value overestimation (see \cref{fig:value_overestimation}) and completely fail to solve the task.
This is because a wider action space makes the critic more vulnerable to function approximation error \citep{fujimoto2018addressing} and the actor excessively maximizes value functions by exploiting this estimation error.
To further support this, in \cref{fig:no_op_actions}, we report additional toy experiments where we introduce redundant no-op actions for training TD3 agents on \texttt{Cheetah Run} task \citep{tassa2020dm_control}.
Here, we find that actor-critic algorithms are indeed vulnerable to value overestimation with high-dimensional action spaces.

\newlength{\panelw}
\setlength{\panelw}{\dimexpr\linewidth/3 - 0.3pt\relax}  
\newlength{\panelh}\setlength{\panelh}{1.50in} 

\begin{figure*}[t!]
\vspace{-0.2in}
\centering
\setlength{\tabcolsep}{0pt} 
\begin{tabular}{@{}p{\panelw} p{\panelw} p{\panelw}@{}}

\subfloat[Return-to-go prediction]{%
  \begin{minipage}[t][\panelh][t]{\panelw}
    \centering
    \includegraphics[width=\linewidth]{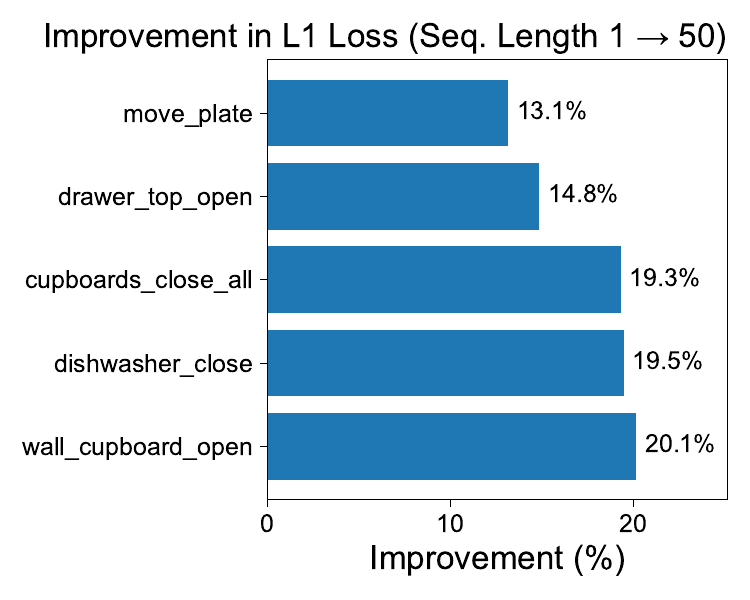}
  \end{minipage}
  \label{fig:return_to_go_prediction}
}
&
\subfloat[Value overestimation on \texttt{stand}]{%
  \begin{minipage}[t][\panelh][t]{\panelw}
    \centering
    \includegraphics[width=\linewidth]{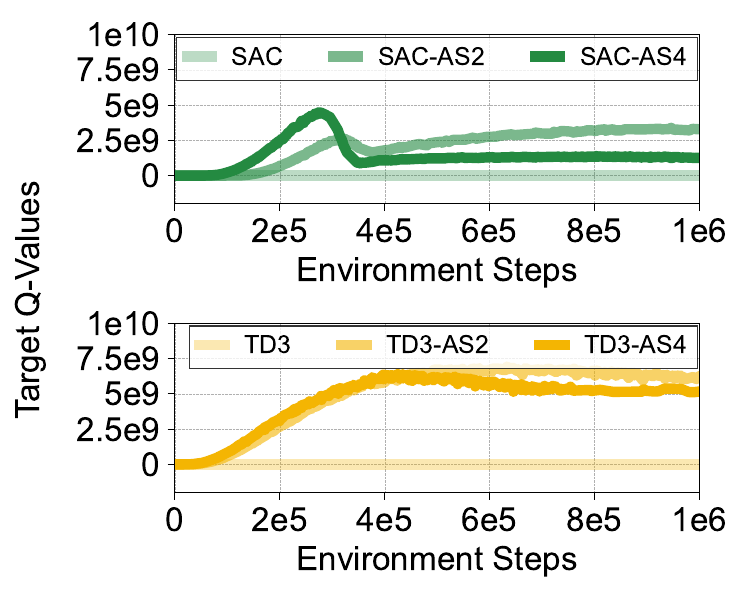}
  \end{minipage}
  \label{fig:value_overestimation}
}
&
\subfloat[Effect of no-op actions]{%
  \begin{minipage}[t][\panelh][c]{\panelw}
    \centering
    \setlength{\tabcolsep}{2pt}
    \renewcommand{\arraystretch}{0.95}
    \resizebox{0.875\linewidth}{!}{%
      \begin{tabular}{ccccc}
        \toprule
        & \multicolumn{2}{c}{TD3} & \multicolumn{2}{c}{CQN} \\
        \cmidrule(lr){2-3}\cmidrule(lr){4-5}
        \makecell[c]{No-op\\added} & \makecell[c]{Return} & \makecell[c]{Value} & \makecell[c]{Return} & \makecell[c]{Value} \\
        \midrule
        0    & 140.7 &  16.25 & 236.6 & 18.8 \\
        54   &   7.18 & -17.45 & 219.5 & 20.3 \\
        144  &   0.56 &  1E8 & 185.4 & 16.4 \\
        294  &   0.27 &  4E8 & 202.9 & 18.7 \\
        \bottomrule
      \end{tabular}%
    }
  \end{minipage}
  \label{fig:no_op_actions}
}
\\
\end{tabular}
\caption{\textbf{Analyses.} (a) We measure the improvement in the validation L1 loss of the return-to-go prediction model with different action sequence lengths. We find that using action sequence of length 50 results in the lower loss than using single-step action.
(b) We find that SAC and TD3 with action sequences suffer from severe value overestimation in \texttt{stand} task from HumanoidBench, which leads to random near-zero performance. (c) Actor-critic algorithms like TD3 become vulnerable to value overestimation when redundant no-op actions are added to the action space. In contrast, a critic-only algorithm that uses discrete actions, CQN, is robust with high-dimensional action spaces.
}
\label{fig:analysis}
\vspace{-0.1in}
\end{figure*}

\begin{figure*}[t!]
\subfloat[Coarse-to-fine inference procedure]
{
\includegraphics[width=0.54\linewidth]{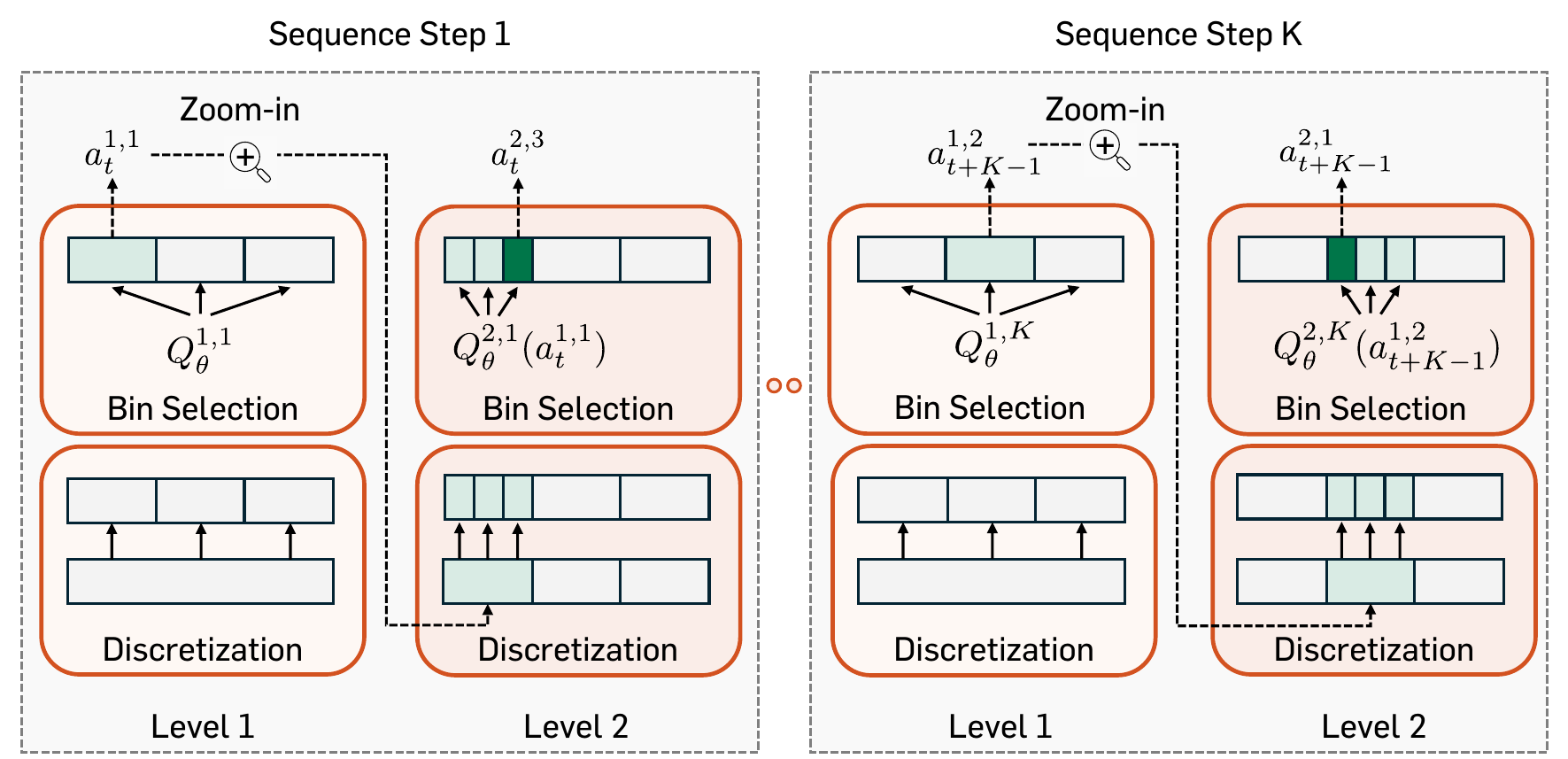}
\label{fig:cqn_as_overview_inference}
}
\subfloat[Architecture]
{
\includegraphics[width=0.41\linewidth]{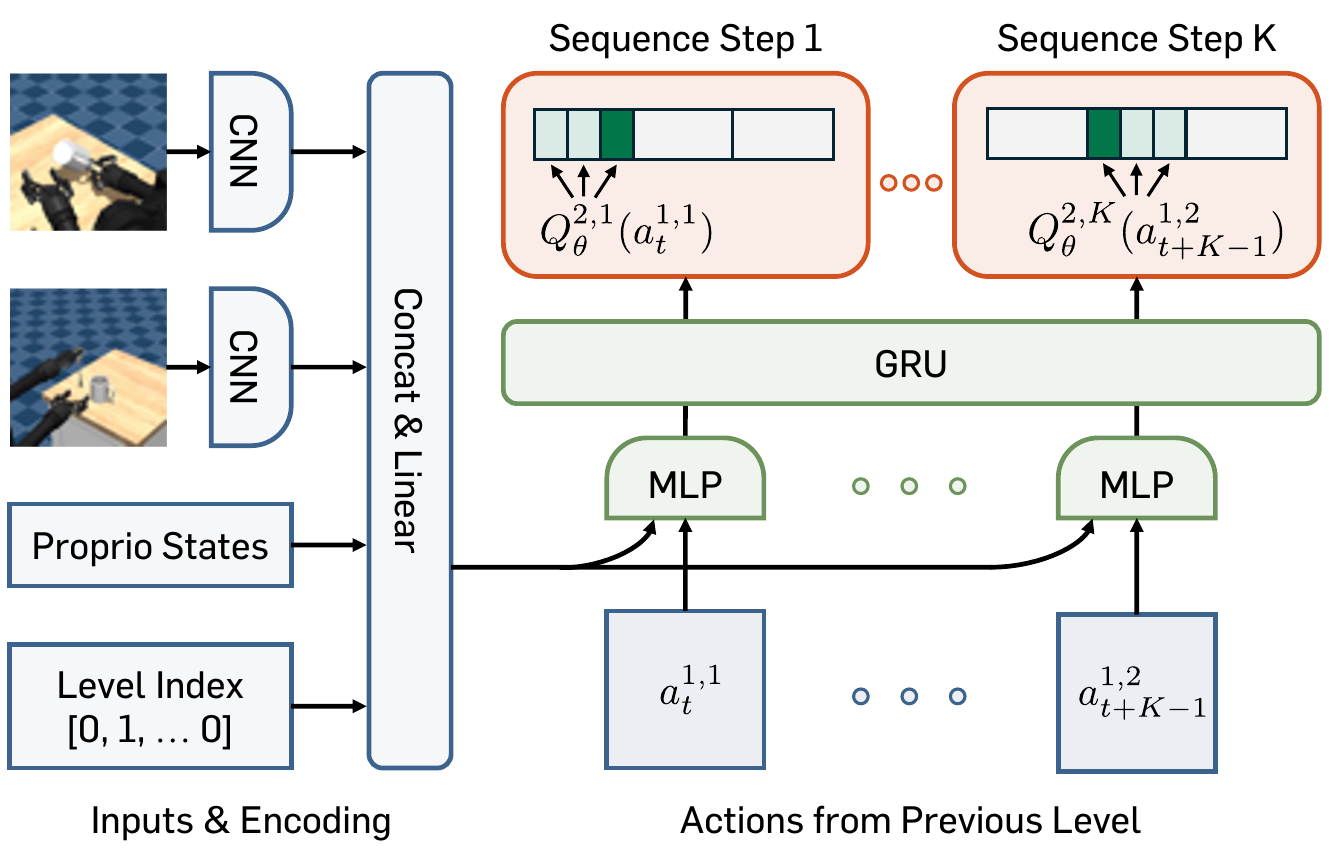}
\label{fig:cqn_as_overview_architecture}
}
\caption{\textbf{Coarse-to-Fine Q-Network with Action Sequence.}
CQN-\textbf{AS} extends Coarse-to-Fine Q-Network (CQN; \citealt{seo2024continuous}), a critic-only RL algorithm for continuous control using discretized actions.
(a) CQN progressively zooms into the action space by discretizing it into $B$ bins and finding the bin with the highest Q-value to further discretize at the next level.
Last level's action sequence is used for controlling robots.
CQN-\textbf{AS} generalizes this to action sequences by computing all $K$ actions in parallel.
(b) We train a critic to predict Q-values over entire action sequences by extracting per-step features and aggregating them with a recurrent network before projection to Q-values.
}
\label{fig:cqn_as_overview}
\end{figure*}

This result motivates us to design our RL algorithm with action sequence upon a recent critic-only algorithm, i.e., Coarse-to-fine Q-Network (CQN; \citealt{seo2024continuous}), which solves continuous control tasks with discrete actions.
Because there is no separate actor that may exploit value functions, i.e., CQN simply selects discrete actions with the highest Q-values, we find that training with action sequences is stable and thus avoids value overestimation problem (see \cref{fig:no_op_actions}).
In particular, we introduce Coarse-to-fine Q-Network with \textbf{A}ction \textbf{S}equence (CQN-\textbf{AS}), which learns a critic network that outputs Q-values over a sequence of actions (see \cref{fig:cqn_as_overview}).
By training the critic network to explicitly learn the consequence of taking a sequence of current and future actions, CQN-\textbf{AS} enables the RL agents to effectively learn useful value functions on challenging robotic tasks.

Our experiments show that CQN-\textbf{AS} improves the performance of CQN on sparse-reward humanoid control tasks from BiGym benchmark \citep{chernyadev2024bigym} that provides human-collected demonstrations and sparse-reward tabletop manipulation tasks from RLBench \citep{james2020rlbench} that provide demonstrations generated via motion-planning.
Considering that CQN-\textbf{AS} is a critic-only algorithm that selects actions with the highest Q-value without a separate actor network, these results highlight the benefit of using action sequences in value learning.

Our contributions can be summarized as below:
\begin{itemize}
    \item We make an observation that shows using action sequences can be useful for RL by enhancing value learning. We also show that standard actor-critic algorithms \citep{haarnoja2018soft,fujimoto2018addressing} suffer from value overestimation when trained with action sequences.
    \item We introduce Coarse-to-fine Q-Network with Action Sequence (CQN-\textbf{AS}) that trains a critic network to output Q-values over action sequences.
    This critic-only algorithm successfully avoids value overestimation problem and enhances the base CQN algorithm.
    \item In a demo-driven RL setup that initializes training with expert demonstrations, we show that CQN-\textbf{AS} surpasses the performance of ACT \citep{zhao2023learning} -- a BC algorithm that trains a Transformer \citep{vaswani2017attention} to predict action sequences.
\end{itemize}

\section{Preliminaries}
\label{sec:preliminaries}

We formulate a robotic control problem as a partially observable Markov decision process \citep{kaelbling1998planning,sutton2018reinforcement}.
At time step $t$, an RL agent encounters an observation $\mathbf{o}_{t}$, executes an action $a_{t}$, receives a reward $r_{t+1}$, and encounters a new observation $\mathbf{o}_{t+1}$ from the environment.
We aim to train a policy $\pi$ that maximizes the expected sum of rewards through RL while using as few online samples as possible, optionally with access to a modest amount of expert demonstrations.

\vspace{-0.025in}
\paragraph{Inputs and encoding}
Given visual observations $\mathbf{o}^{v}_{t} = \{\mathbf{o}^{v_{1}}_{t}, ..., \mathbf{o}^{v_{M}}_{t}\}$ from $M$ cameras, we encode each $\mathbf{o}^{v_{i}}_{t}$ using convolutional neural networks (CNN) into $\mathbf{h}_{t}^{v_{i}}$.
We then process them through a series of linear layers to fuse them into $\mathbf{h}_{t}^{v}$.
If low-dimensional observations $\mathbf{o}_{t}^{\tt{low}}$ are available along with visual observations, we process them through a series of linear layers to obtain $\mathbf{h}_{t}^{\tt{low}}$.
We then use concatenated features $\mathbf{h}_{t} = [\mathbf{h}^{v}_{t}, \mathbf{h}^{\tt{low}}_{t}]$ as inputs to the critic network.
In domains without vision sensors, we simply use $\mathbf{o}_{t}^{\tt{low}}$ as $\mathbf{h}_{t}$ without encoding the low-dimensional observations.

\vspace{-0.025in}
\paragraph{Coarse-to-fine Q-Network}
Coarse-to-fine Q-Network (CQN; \citealt{seo2024continuous}) is a critic-only RL algorithm that solves continuous control tasks with discrete actions.
CQN trains an RL agent to learn to select coarse discrete actions in shallower levels with larger bin sizes, and then refine their choices by selecting finer-grained actions in deeper levels with smaller bin sizes.
Specifically, CQN iterates the procedures of (i) discretizing the continuous action space into multiple bins and (ii) selecting the bin with the highest Q-value to further discretize.
This reformulates the continuous control problem as a multi-level discrete control problem, allowing for the use of ideas from sample-efficient discrete RL algorithms \citep{mnih2015human,silver2017mastering} for continuous control.

Formally, let $a_{t}^{l}$ be an action at level $l$ with $a_{t}^{0}$ being the zero vector.\footnote{For simplicity, we describe CQN and CQN-\textbf{AS} with a single-dimensional action in the main section. See \cref{appendix:full_description_cqn_as} for full description with $N$-dimensional actions.}
We then define the coarse-to-fine critic to consist of multiple Q-networks which compute Q-values for actions at each level $a_{t}^{l}$, given the features $\mathbf{h}_{t}$ and actions from the previous level $a_{t}^{l-1}$, as follows:
\begin{equation}
Q^l_\theta(\mathbf{h}_t, a^{l-1}_t) = \left[Q^l_\theta(\mathbf{h}_t, a^l_t = a^{l,b}_t, a^{l-1}_t) \right]_{b=1}^{B} \in \mathbb{R}^B
\end{equation}
where $a_{t}^{l,b}$ denotes an action for each bin $b$ and $B$ is the number of bins for each level.
We note that CQN uses scalar values representing the center of each bin for previous level's action $a^{l-1}_{t}$, enabling the network to locate itself without access to all previous levels' actions.
We optimize each Q-network at level $l$ with the following objective:
\vspace{-0.0in}
\begin{gather*}
\begin{aligned}
    \mathcal{L}^{l} = \big( Q_{\theta}^{l}(\mathbf{h}_{t}, a_{t}^{l}, a_{t}^{l-1}) - r_{t+1} - \gamma \max_{a'}Q_{\bar{\theta}}^{l}(\mathbf{h}_{t+1}, a', \pi^{l-1}(\mathbf{h}_{t+1}) \big),
\label{eq:cqn_objective}
\end{aligned}
\end{gather*}
where $\bar{\theta}$ are delayed parameters for a target network \citep{polyak1992acceleration} and $\pi^{l}$ is a policy that outputs the action $a_{t}^{l}$ at each level $l$ via the inference steps with our critic, \textit{i.e.,} $\pi^{l}(\mathbf{h}_{t}) = a_{t}^{l}$.
Specifically, to output actions at time step $t$, CQN first initializes constants $a_{t}^{\tt{low}}$ and $a_{t}^{\tt{high}}$ with $-1$ and $1$.
Then the following steps are repeated for $l \in \{1, ..., L\}$:
\begin{itemize}
[leftmargin=20pt]
    \item  [$\bullet$] Step 1 (Discretization): Discretize an interval $[a_{t}^{\tt{low}}, a_{t}^{\tt{high}}]$ into $B$ uniform intervals, and each of these intervals become an action space for $Q_{\theta}^{l}$.
    \item  [$\bullet$] Step 2 (Bin selection): Find a bin with the highest Q-value and set $a_{t}^{l}$ to the centroid of the bin.
    \item  [$\bullet$] Step 3 (Zoom-in): Set $a_{t}^{\tt{low}}$ and $a_{t}^{\tt{high}}$ to the minimum and maximum of the selected bin, which intuitively can be seen as zooming-into each bin.
\end{itemize}
We then use the last level's action $a_{t}^{L}$ as the action at time step $t$.
For more details, including the inference procedure for computing Q-values, we refer readers to \cref{appendix:full_description_cqn_as}.

\section{Method}
We present Coarse-to-fine Q-Network with \textbf{A}ction \textbf{S}equence (CQN-\textbf{AS}), a value-based RL algorithm that learns a critic network that outputs Q-values for \textit{a sequence of actions} $a_{t:t+K} = \{a_{t}, ..., a_{t+K-1}\}$ for a given observation $\mathbf{o}_{t}$.
Our main motivation comes from one of the key ideas in recent BC approaches: predicting \textit{action sequences}, which helps resolve ambiguity when approximating noisy distributions of expert demonstrations \citep{zhao2023learning,chi2023diffusion}.
Similarly, by explicitly learning Q-values of a sequence of actions from the given state, our approach mitigates the challenge of learning Q-values with noisy trajectories.
We provide the overview of CQN-\textbf{AS} in \cref{fig:cqn_as_overview}.

\subsection{Coarse-to-fine Critic with Action Sequence}
\label{sec:c2f_critic_as}

\paragraph{Objective}
Let $a_{t:t+K}^{l} = \{ a_{t}^{l}, ..., a_{t+K-1}^{l}\}$ be an action sequence at level $l$ and $a_{t:t+K}^{0}$ be a zero vector.
We design our coarse-to-fine critic network to consist of multiple Q-networks that compute Q-values for each action at sequence step $k \in \{1, ..., K\}$ and level $l \in \{1, ..., L\}$:
\vspace{-0.025in}
\begin{align*}
    Q_{\theta}^{l, k}(\mathbf{h}_{t}, a_{t:t+K}^{l-1}) = \left[Q_{\theta}^{l, k}(\mathbf{h}_{t}, a_{t+k-1}^{l,b}, a_{t:t+K}^{l-1})\right]_{b=1}^{B} \in \mathbb{R}^B
\end{align*}
where $a_{t+k-1}^{l,b}$ denotes an action for each bin $b$ at step $k$. We optimize our critic network with the following objective:
\begin{gather}
\begin{aligned}
    \sum\nolimits_{k}\sum\nolimits_{l}\big(  Q_{\theta}^{l,k}(\mathbf{h}_{t}, a_{t+k-1}^{l}, a_{t:t+K}^{l-1}) - \sum\nolimits_{i=1}^{N} r_{t+i} - \gamma \max_{a'}Q_{\bar{\theta}}^{l,k}(\mathbf{h}_{t+1}, a', \pi^{l-1}_{K}(\mathbf{h}_{t+1}) \big)^{2},
\end{aligned}
\label{eq:cqn_as_objective}
\end{gather}
where $N$ is a hyperparameter for $N$-step return and $\pi_{K}^{l}$ is an action sequence policy that outputs the action sequence $\mathbf{a}_{t:t+K}^{l}$ by following the similar inference procedure as in \cref{sec:preliminaries} (see \cref{fig:cqn_as_overview_inference}).
In practice, we compute Q-values for all sequence step $k \in \{1, ..., K\}$ in parallel, which is possible as Q-values for future actions depend only on features $\mathbf{h}_{t}$ but not on previous actions.

\paragraph{Remarks on objective with $N$-step return}
We note that any $N$-step return can be used in \cref{eq:cqn_as_objective} because the network can learn the long-term value of outputting action $a_{t+k}$ from bootstrapping.
There is a trade-off: if one considers a short $N$-step return, it can cause a challenge as the setup becomes a delayed reward setup; but training with higher $N$-step return may introduce variance \citep{sutton2018reinforcement}.
In our considered setups, we empirically find that using common values $N \in \{1, 4\}$ works the best. We provide empirical analysis on the effect of $N$ in \cref{fig:effect_of_nstep}.

\vspace{-0.06in}
\paragraph{Architecture}
Our critic network initially extracts features for each sequence step $k$ and aggregates features from multiple steps with a recurrent network (see \cref{fig:cqn_as_overview_architecture}).
This architecture is helpful in cases where a single-step action is already high-dimensional so that concatenating them make inputs too high-dimensional.
Specifically, let $\mathbf{e}_{k}$ denote an one-hot encoding for $k$.
At each level $l$, we construct features for each sequence step $k$ as $\mathbf{h}^{l}_{t,k} = \left[\mathbf{h}_{t}, a_{t+k-1}^{l-1}, \mathbf{e}_{k}\right]$.
We then encode each $\mathbf{h}_{t,k}^{l}$ with a shared MLP network and process them through GRU \citep{cho2014learning} to obtain $\mathbf{s}^{l}_{t,k} = f^{\tt{GRU}}_{\theta}(f^{\tt{MLP}}_{\theta}(\mathbf{h}_{t,1}^{l}), ..., f^{\tt{MLP}}_{\theta}(\mathbf{h}_{t,k}^{l})$.
We find that this design empirically performs better than directly giving actions as inputs to GRU.
We then use a shared projection layer to map each $\mathbf{s}_{t,k}^{l}$ into Q-values at each sequence step $k$, \textit{i.e.,} $[Q_{\theta}^{l, k}(\mathbf{h}_{t}, a_{t+k-1}^{l,b}, a_{t:t+K}^{l-1})]_{b=1}^{B} = f_{\theta}^{\tt{proj}}(\mathbf{s}_{t,k}^{l})$.

\subsection{Action Execution and Training Details}
\label{sec:controlling_robots}

\paragraph{Executing action with temporal ensemble}
With the policy that outputs an action sequence $a_{t:t+K}$,
one question is how to execute actions at time step $i \in \{t, ..., t+K-1\}$.
For this, we use \textit{temporal ensemble} \citep{zhao2023learning} that computes $a_{t:t+K}$ every time step, saves it to a buffer, and executes a weighted average $\sum_{i} w_{i} \bar{a}_{t}^{i} / \sum w_{i}$ where $\bar{a}_{t}^{i}$ denotes an action for step $t$ computed at step $t-i$, $w_{i} = \exp (-m * i)$ denotes a weight that assigns higher value to more recent actions, with $m$ as a hyperparameter that adjusts the weighting magnitude.
We find this scheme outperforms the alternative of computing $a_{t:t+K}$ every $K$ steps and executing each action for subsequent $K$ steps on most tasks we considered, except on several tasks that need reactive control.

\vspace{-0.06in}
\paragraph{Storing training data}
When storing samples from the environment, we store a transition $(\mathbf{o}_{t}, \hat{a}_{t}, r_{t+1}, \mathbf{o}_{t+1})$ where $\hat{a}_{t}$ denotes an action executed at time step $t$.
For instance, if we use temporal ensemble for action execution, $\hat{a}_{t}$ is a weighted average of action outputs obtained from previous $K$ time steps, i.e., $\hat{a}_{t} = \sum_{i} w_{i} \bar{a}_{t}^{i} / \sum w_{i}$.

\vspace{-0.06in}
\paragraph{Sampling training data from a replay buffer}
When sampling training data from the replay buffer, we sample a transition with action sequence, \textit{i.e.,} $(\mathbf{o}_{t}, \hat{a}_{t:t+K}, r_{t+1}, \mathbf{o}_{t+1})$.
If we sample time step $t$ near the end of episode so that we do not have enough data to construct a full action sequence, we fill the action sequence with \textit{null} actions.
In particular, in position control where we specify the position of joints or end effectors, we repeat the action from the last step so that the agent learns not to change the position.
In torque control where we specify the force to apply, we set the action after the last step to zero so that agent learns to not to apply force.

\begin{figure*}[t!]
    \centering
    \includegraphics[width=0.99\textwidth]{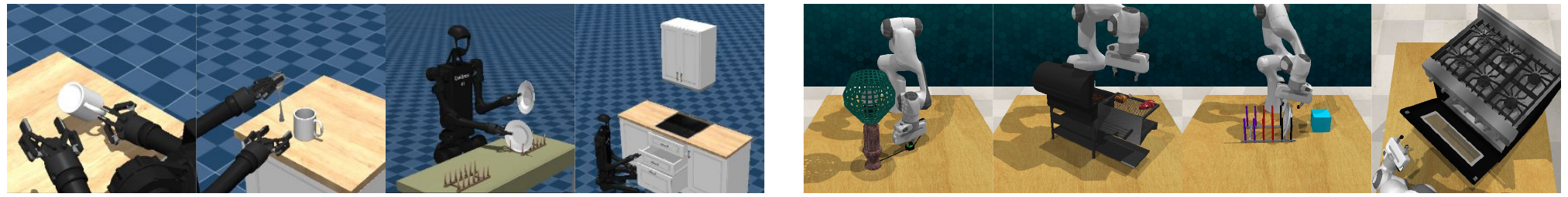}
    \caption{\textbf{Examples of robotic tasks.} We study CQN-\textbf{AS} on 25 humanoid control tasks from BiGym \citep{chernyadev2024bigym} and 20 tabletop manipulation tasks from RLBench \citep{james2020rlbench}.}
    \label{fig:tasks}
\end{figure*}

\section{Experiment}
\label{sec:experiment}
We study CQN-\textbf{AS} on 25 humanoid control tasks from BiGym \citep{chernyadev2024bigym} and 20 tabletop manipulation tasks from RLBench \citep{james2020rlbench} (see \cref{fig:tasks} for examples of robotic tasks).
To focus on challenging robotic tasks that aim to induce policies generating realistic behaviors, we consider a practical setup of demo-driven RL where we initialize training with a modest amount of expert demonstrations and then train with online data.
In particular, our experiments are designed to investigate the following questions:
\begin{itemize}
    \item [$\bullet$] Can CQN-\textbf{AS} quickly match the performance of a recent BC algorithm \citep{zhao2023learning} and surpass it through online learning? How does CQN-\textbf{AS} compare to previous model-free RL algorithms \citep{yarats2022mastering,seo2024continuous}? 
    \item [$\bullet$] What is the effect of each component in CQN-\textbf{AS}?
    \item [$\bullet$] Under which conditions is CQN-\textbf{AS} effective?
\end{itemize}

\vspace{-0.06in}
\paragraph{Baselines}
We consider model-free RL baselines that learn deterministic policies, as we find that stochastic policies struggle to solve fine-grained manipulation tasks.
Specifically, we consider (i) Coarse-to-fine Q-Network (CQN; \citealt{seo2024continuous}), our backbone algorithm and (ii) DrQ-v2+, an optimized demo-driven variant of an actor-critic algorithm DrQ-v2 \citep{yarats2022mastering} that uses a deterministic policy algorithm and data augmentation.
We further consider (iii) Action Chunking Transformer (ACT; \citealt{zhao2023learning}) that trains a transformer \citep{vaswani2017attention} policy to predict action sequences from expert demonstrations and utilizes temporal ensemble, as our BC baseline.

\paragraph{Implementation details}
For training with expert demonstrations, we follow the setup of \citet{seo2024continuous}.
We keep a separate replay buffer that stores demonstrations and sample half of training data from demonstrations.
We also relabel successful online episodes as demonstrations and store them in the demonstration replay buffer.
For CQN-\textbf{AS}, we use an auxiliary BC loss from \citet{seo2024continuous} based on large margin loss \citep{hester2018deep}.
For actor-critic baselines, we use an auxiliary BC loss that minimizes L2 loss between the policy outputs and expert actions.

\begin{figure*}[t!]
    \centering
    \includegraphics[width=0.99\textwidth]{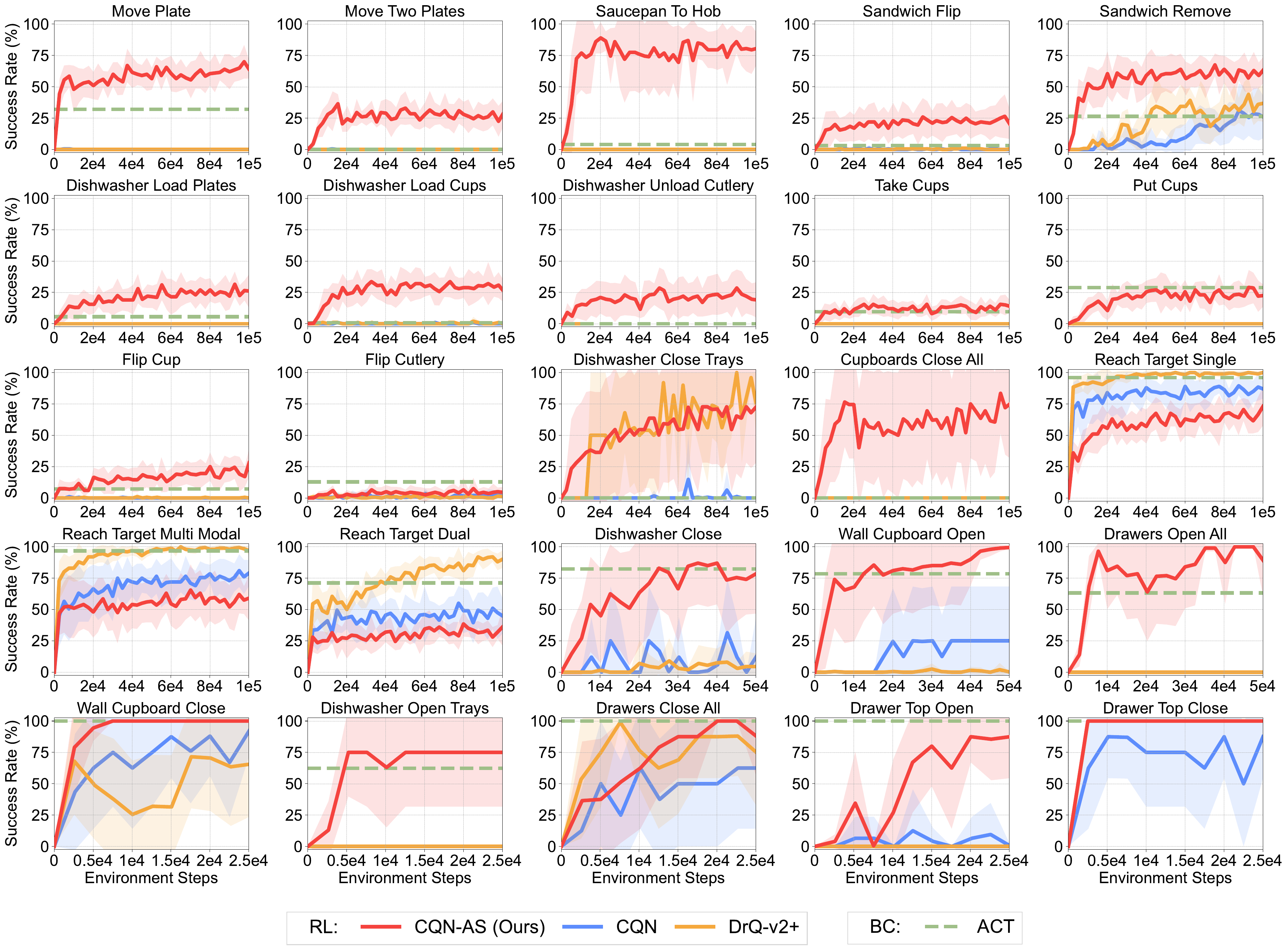}
    \caption{\textbf{BiGym results} on 25 sparsely-rewarded mobile bi-manual manipulation tasks.
    All RL algorithms are trained from scratch, with a replay buffer initialized with 17 to 60 \textit{human-collected} demonstrations, and with an auxiliary BC objective.
    We report the success rate over 25 episodes.
    The solid line and shaded regions represent the mean and confidence intervals, respectively, across 8 runs.
    }
    \label{fig:bigym_experiments}
    \vspace{-0.125in}
    \end{figure*}

\vspace{-0.04in}
\subsection{BiGym Experiments}
\label{sec:bigym_experiments}
We study CQN-\textbf{AS} on mobile bi-manual manipulation tasks from BiGym \citep{chernyadev2024bigym}.
BiGym's \textit{human-collected} demonstrations are often noisy and multi-modal, posing challenges for RL algorithms. These algorithms must effectively leverage the information within demonstrations to learn strong initial behaviors, thereby mitigating exploration difficulties in sparsely rewarded tasks.

\vspace{-0.04in}
\paragraph{Setup}
We consider 25 BiGym tasks with 17 to 60 demonstrations\footnote{BiGym benchmark provides different number of successful demonstrations for each task. But we use the same number of demonstrations for all algorithms. See \cref{appendix:experimental_details} for more details.}.
We use RGB observations with 84$\times$84 resolution from \texttt{head}, \texttt{left\_wrist}, and \texttt{right\_wrist} cameras.
We also use low-dimensional proprioceptive states.
We use (i) absolute joint position control action mode and (ii) floating base that replaces locomotion with classic controllers.
We use the same set of hyperparameters for all the tasks.
Details on BiGym experiments are available in \cref{appendix:experimental_details}.

\vspace{-0.06in}
\paragraph{Comparison to baselines}
\cref{fig:bigym_experiments} shows that CQN-\textbf{AS} quickly matches the performance of ACT and outperforms it through online learning on most tasks, while other RL algorithms fail to do so especially on challenging long-horizon tasks such as \texttt{Move} \texttt{Plate} and \texttt{Saucepan} \texttt{To} \texttt{Hob}.
A notable result here is that CQN-\textbf{AS} \textit{enables} solving challenging BiGym tasks while other RL baselines completely fail as they achieve a 0\% success rate on many tasks.

\vspace{-0.06in}
\paragraph{Limitation}
However, CQN-\textbf{AS} struggles to achieve meaningful success rate on some of the long-horizon tasks that require interaction with delicate objects such as cups or cutlery.
This leaves room for future work to incorporate advanced vision encoders \citep{he2016deep} or critic architectures \citep{chebotar2023q,springenberg2024offline}.

\begin{figure*}[t!]
\centering
\includegraphics[width=0.99\textwidth]{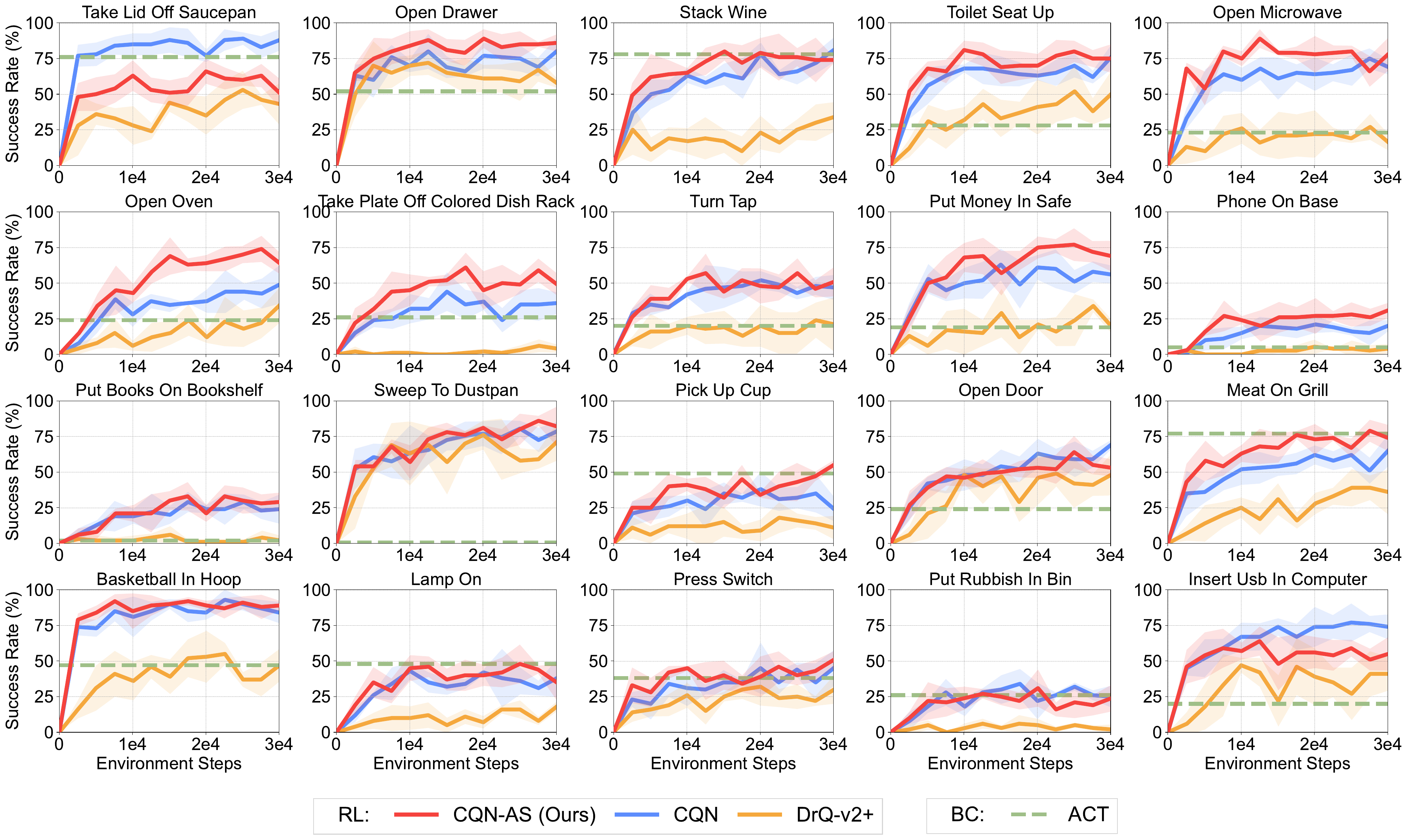}
\caption{\textbf{RLBench results} on 20 sparsely-rewarded tabletop manipulation tasks from RLBench \citep{james2020rlbench}. 
All RL algorithms are trained from scratch, with a replay buffer initialized with 100 \textit{synthetic} demonstrations generated via motion-planning, and with an auxiliary BC objective.
\textit{As expected}, with synthetic demonstrations, CQN-\textbf{AS} achieves similar performance to CQN on most tasks.
However, CQN-\textbf{AS} often significantly outperforms baselines on several challenging, long-horizon tasks such as \texttt{Open} \texttt{Oven}.
We report the success rate over 25 episodes.
    The solid line and shaded regions represent the mean and confidence intervals, respectively, across 4 runs.}
\label{fig:rlbench_experiments}
\end{figure*}

\subsection{RLBench Experiments}
We also study CQN-\textbf{AS} on manipulation tasks from RLBench \citep{james2020rlbench}.  
Unlike BiGym, RLBench provides synthetic demonstrations generated via motion planning, which are cleaner and more consistent.  
This allows us to examine whether CQN-\textbf{AS} is also effective in settings with \textit{clean}, unambiguous demonstrations -- where the effect of each single-step action is easier to interpret.

\vspace{-0.025in}
\paragraph{Setup}
We use the official CQN implementation for collecting demonstrations and reproducing the baseline results on the same set of tasks.
We use RGB observations with 84$\times$84 resolution from \texttt{front}, \texttt{wrist}, \texttt{left\_shoulder}, and \texttt{right\_shoulder} cameras.
We also use low-dimensional proprioceptive states consisting of 7-dimensional joint positions and a binary value for gripper open.
We use 100 demonstrations and delta joint position control action mode.
We use the same set of hyperparameters for all the tasks, in particular, we use action sequence of length 4.
More details on RLBench experiments are available in \cref{appendix:experimental_details}.

\vspace{-0.025in}
\paragraph{CQN-AS is also effective with \textit{clean} demonstrations}
Because RLBench provides synthetic \textit{clean} demonstrations, as we expected, \cref{fig:rlbench_experiments} shows that CQN-\textbf{AS} achieves \textit{similar} performance to CQN on most tasks, except 2/25 tasks where it hurts the performance.
But we still find that CQN-\textbf{AS} achieves quite superior performance to CQN on some challenging long-horizon tasks such as \texttt{Open} \texttt{Oven} or \texttt{Take} \texttt{Plate} \texttt{Off} \texttt{Colored} \texttt{Dish} \texttt{Rack}.
These results show that CQN-\textbf{AS} can be used in various benchmark with different characteristics.

\begin{figure*}[t]
    \centering
    \includegraphics[width=0.99\textwidth]{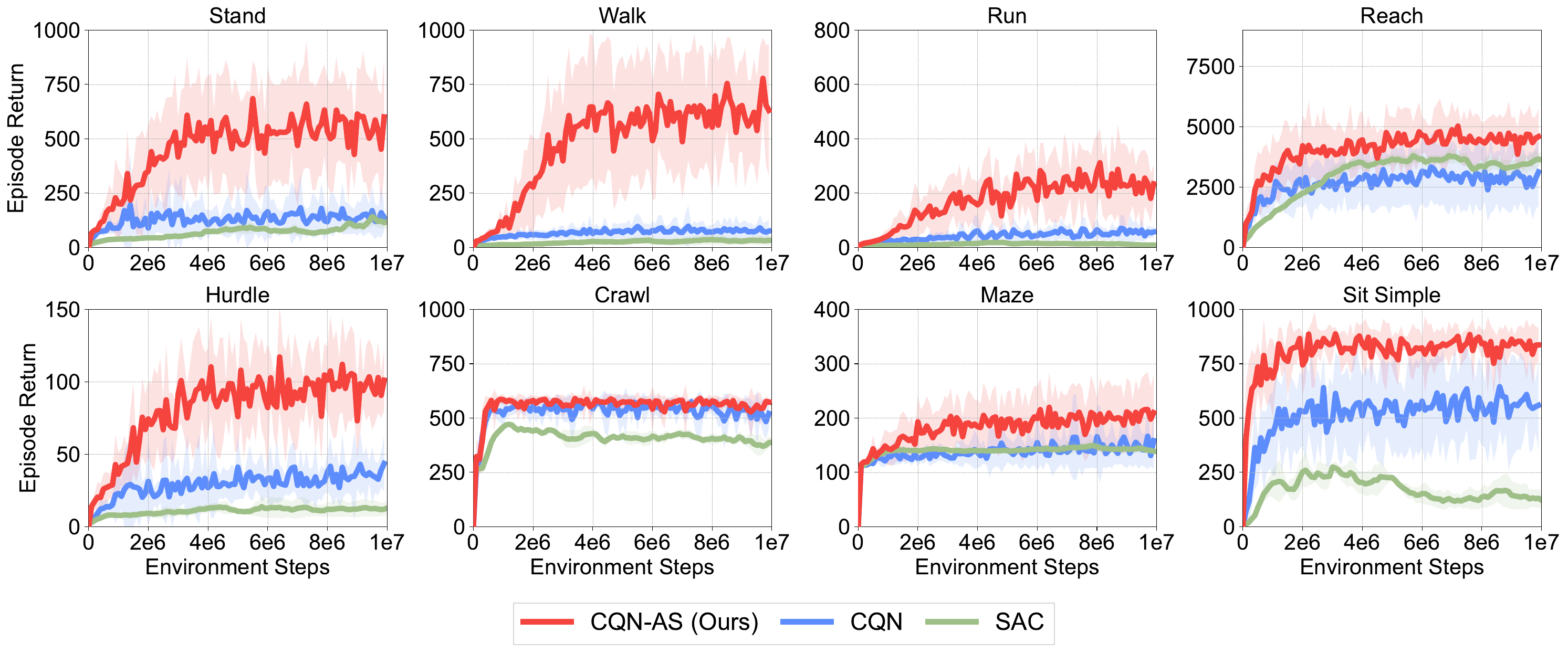}
    \caption{\textbf{HumanoidBench results} on eight densely-rewarded humanoid control tasks \citep{sferrazza2024humanoidbench}.
    All the experiments start from scratch and all the methods do not have an auxiliary BC objective.
    CQN-\textbf{AS} significantly improves the performance of underlying RL algorithm CQN, while outperforming a model-free RL baseline, SAC.
    For CQN-\textbf{AS} and CQN, we report the results aggregated over 8 runs. 
    For SAC, we report the results aggregated over 3 runs available from public website.
    The solid line and shaded regions represent the mean and confidence intervals.}
    \label{fig:humanoidbench_experiments}
\end{figure*}

\begin{figure*}[t!]
    \centering
    \subfloat[Effect of action sequence length]{
    \includegraphics[width=0.465\linewidth]{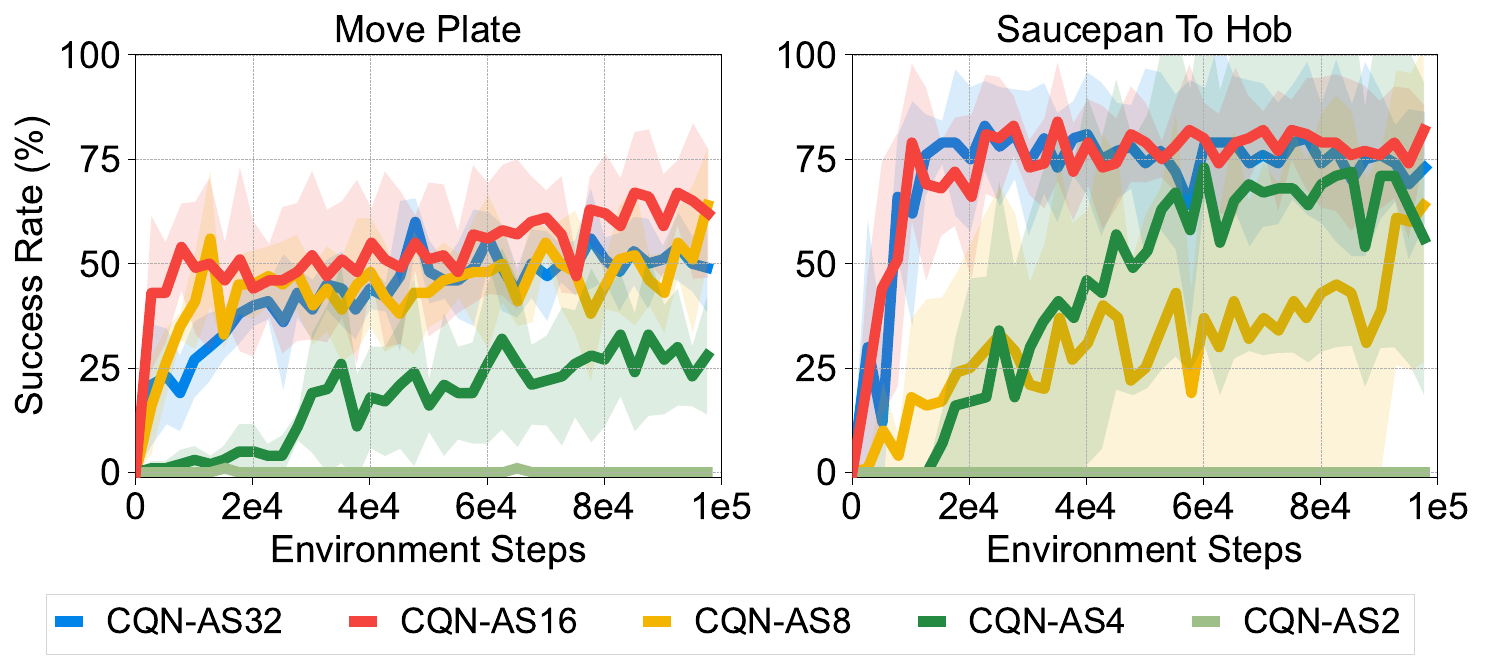}
    \label{fig:effect_of_action_sequence}
    }
    \subfloat[Effect of RL objective]{
    \includegraphics[width=0.465\linewidth]{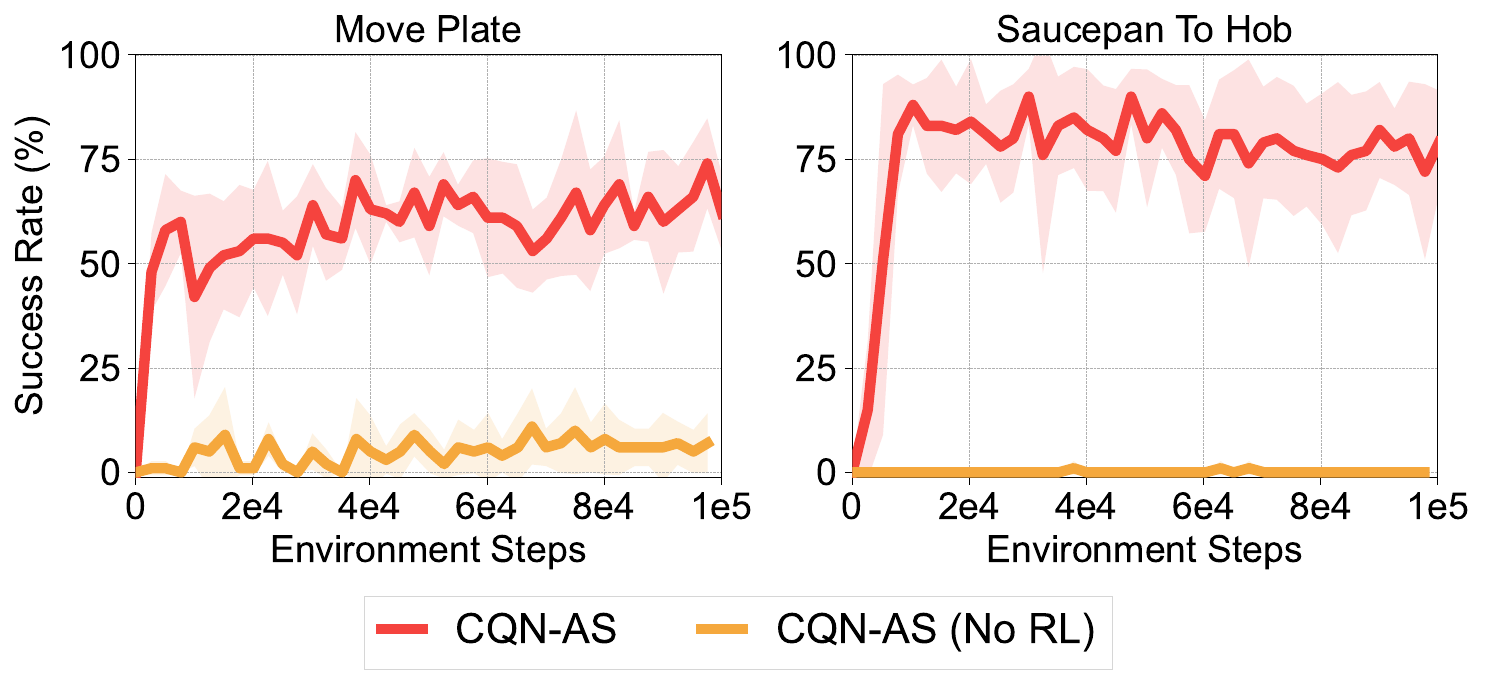}
    \label{fig:effect_of_rl_objective}
    }
    \vspace{-0.1in}
    \\
    \subfloat[Effect of $N$-step]{
    \includegraphics[width=0.465\linewidth]{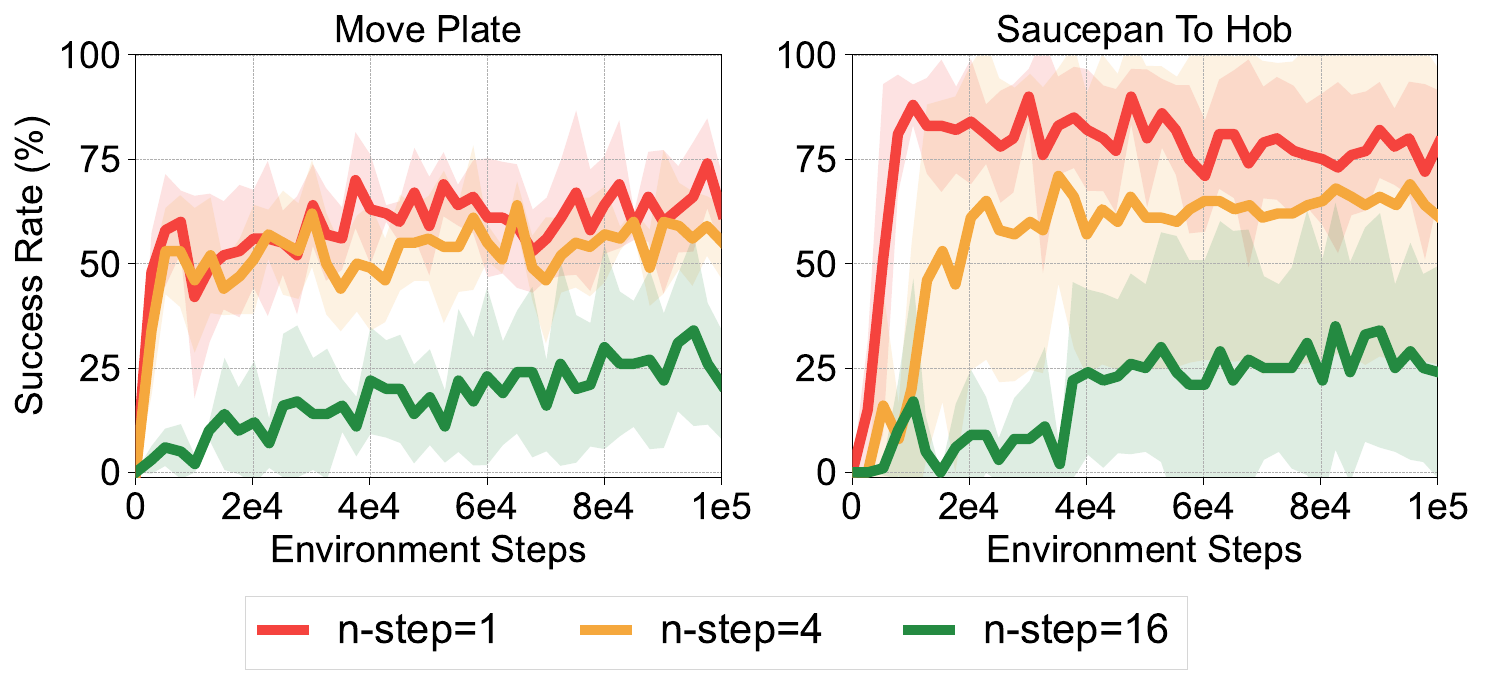}
    \label{fig:effect_of_nstep}
    }
    \subfloat[Effect of temporal ensemble ($m=0.01$)]{
    \includegraphics[width=0.465\linewidth]{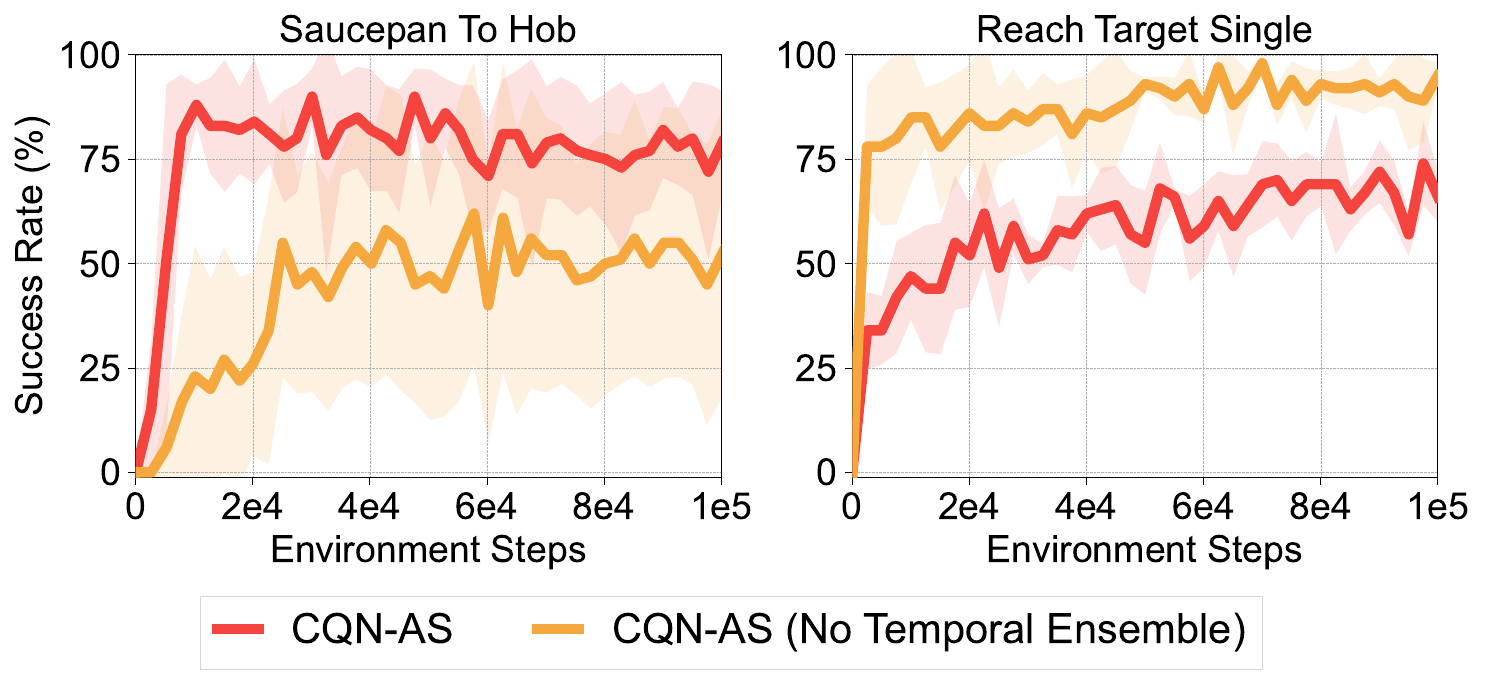}
    \label{fig:effect_of_temporal_ensemble}
    }
    \vspace{-0.1in}
    \\
    \subfloat[Effect of temporal ensemble magnitude $m$]{
    \includegraphics[width=0.465\linewidth]{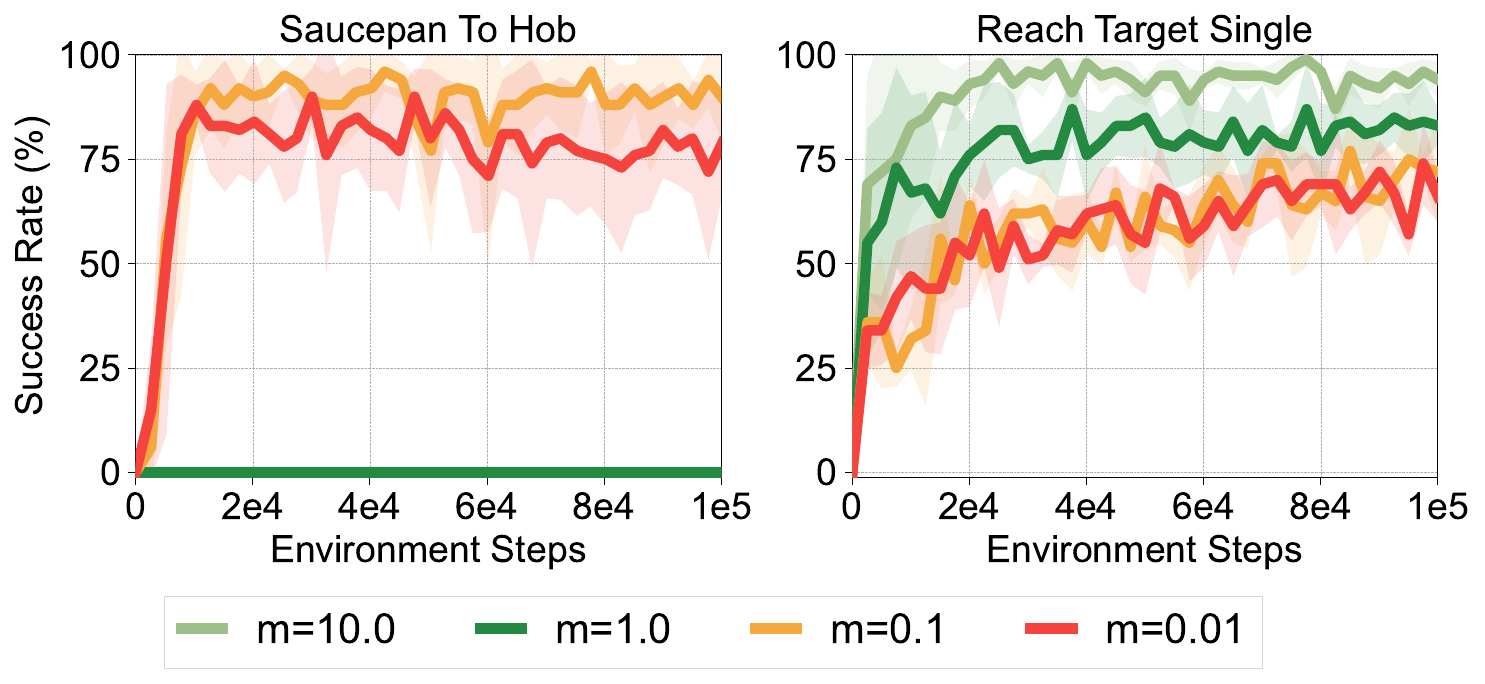}
    \label{fig:effect_of_magnitude_m}
    }
    \subfloat[Failure mode: Torque control]{
    \includegraphics[width=0.465\linewidth]{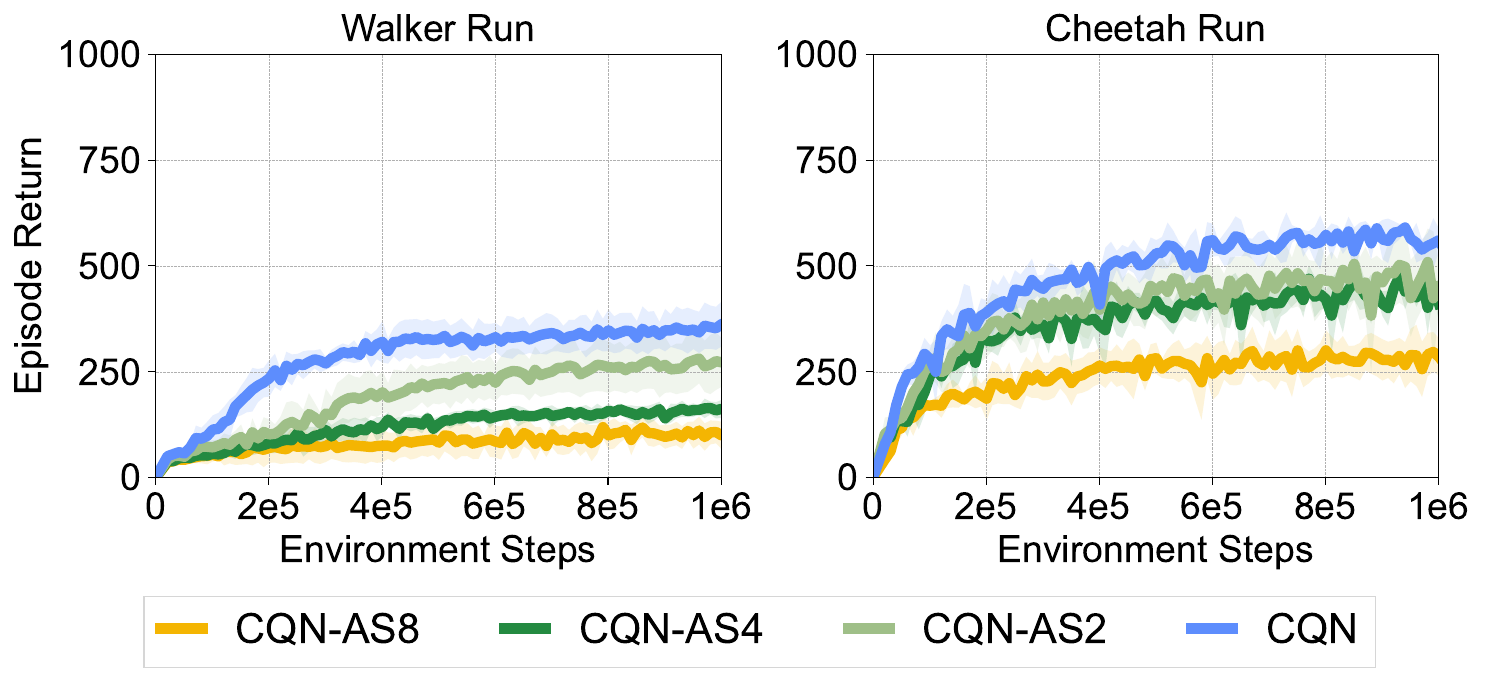}
    \label{fig:failure_mode}
    }
    \caption{\textbf{Ablation studies and analysis} on the effect of (a) action sequence, (b) RL objective, (c) $N$-step return, and (d \& e) temporal ensemble.
    (f) We also provide results on locomotion tasks from DeepMind Control Suite \citep{tassa2020dm_control}, where CQN-\textbf{AS} fails to improve performance.
    The solid line and shaded regions represent the mean and confidence intervals, respectively, across 4 runs.}
    \label{fig:ablation_and_analysis}
    \vspace{-0.1in}
    \end{figure*}

\subsection{HumanoidBench Experiments}
To show that CQN-\textbf{AS} is generally applicable to tasks without demonstrations, we also study CQN-\textbf{AS} on densely-rewarded tasks from HumanoidBench \citep{sferrazza2024humanoidbench}.

\paragraph{Setup}
We follow a standard setup that trains RL agents from scratch.
We use low-dimensional states consisting of proprioception and privileged task information as inputs.
For tasks, we simply select the first 8 locomotion tasks in the benchmark.
For baselines, we consider CQN and Soft Actor-Critic (SAC) \citep{haarnoja2018soft}.
For SAC, we use the results available from HumanoidBench repository, which are evaluated on \textit{tasks with dexterous hands}.
For CQN-\textbf{AS} and CQN, we also evaluate them on tasks with hands.
We use the same set of hyperparameters for all the tasks (see \cref{appendix:experimental_details}).

\paragraph{Comparison to baselines}
\cref{fig:humanoidbench_experiments} shows that, by learning the critic network with action sequence, CQN-\textbf{AS} outperforms other model-free RL baselines, \textit{i.e.,} CQN  and SAC, on most tasks.
In particular, the difference between CQN-\textbf{AS} and baselines becomes larger as the task gets more difficult, \textit{e.g.,} baselines fail to achieve high episode return on \texttt{Walk} and \texttt{Run} tasks but CQN-\textbf{AS} achieves strong performance.
This result shows that our idea of using action sequence can be applicable to generic setup without demonstrations.

\subsection{Ablation Studies, Analysis, Failure Cases}
\label{sec:ablation_analysis_failure}
\paragraph{Effect of action sequence length}
\cref{fig:effect_of_action_sequence} shows the performance of CQN-\textbf{AS} with different action sequence lengths.
We find that training the critic network with longer action sequences tends to consistently improve performance, plateaus or decreases performance if the sequences get too long.

\vspace{-0.075in}
\paragraph{RL objective is crucial for strong performance}
\cref{fig:effect_of_rl_objective} shows the performance of CQN-\textbf{AS} without RL objective that trains the model only with BC objective on successful demonstrations.
We find this baseline significantly underperforms CQN-\textbf{AS}, which shows that RL objective enables the agent to learn from trial-and-error experiences.

\vspace{-0.075in}
\paragraph{Effect of $N$-step return}
\cref{fig:effect_of_nstep} shows experimental results with varying $N$-step returns.
We find that too high $N$-step return significantly degrades performance.
We hypothesize this is because the variance from $N$-step return makes it difficult to learn useful value functions.

\vspace{-0.075in}
\paragraph{Effect of temporal ensemble}
\cref{fig:effect_of_temporal_ensemble} shows that performance degrades without temporal ensemble on \texttt{Saucepan} \texttt{To} \texttt{Hob} as temporal ensemble induces a smooth motion and thus improves performance in fine-grained control tasks.
But we also find that temporal ensemble is harmful on \texttt{Reach} \texttt{Target} \texttt{Single}.
This is because temporal ensemble uses predictions from previous steps and thus makes it difficult to refine behaviors based on recent observations.
Nonetheless, we use temporal ensemble for all the tasks as it helps on most tasks and we aim to use the same set of hyperparameters.

\vspace{-0.075in}
\paragraph{Effect of temporal ensemble magnitude}
We further provide results with different temporal ensemble magnitudes by adjusting a hyperparameter $m$ in \cref{fig:effect_of_magnitude_m}.
Here, higher $m$ puts higher weights on recent actions and thus very high $m$ corresponds to using only first action.
Similarly to previous paragraph, we find that higher $m$ leads to better performance on \texttt{Reach} \texttt{Target} \texttt{Single} that needs fast reaction, but degrades performance on \texttt{Saucepan} \texttt{To} \texttt{Hob}.

\vspace{-0.05in}
\paragraph{Failure case: Torque control}
\cref{fig:failure_mode} shows that CQN-\textbf{AS} underperforms CQN on locomotion tasks with torque control, which are drawn from the DeepMind Control Suite \citep{tassa2020dm_control}.
We hypothesize that this performance degradation arises because sequences of joint positions tend to have clearer semantic structure in joint space, making them easier to learn from compared to sequences of raw torques.
Addressing this failure case represents an interesting direction for future work.

\vspace{-0.05in}
\section{Related Work}
\vspace{-0.05in}

\paragraph{Behavior cloning with action sequence}
Recent behavior cloning approaches have shown that predicting a sequence of actions enables the policy to imitate noisy expert trajectories and helps in dealing with idle actions from human pauses during data collection \citep{zhao2023learning,chi2023diffusion}.
Notably, \citet{zhao2023learning} train a transformer model \citep{vaswani2017attention} that predicts action sequence and \citet{chi2023diffusion} train a denoising diffusion model \citep{ho2020denoising} that approximates the action distributions.
This idea has been extended to multi-task setup \citep{bharadhwaj2024roboagent}, mobile manipulation \citep{fu2024mobile} and humanoid control \citep{fu2024humanplus}.
Our work is inspired by this line of research and proposes to learn RL agents with action sequence.

\vspace{-0.075in}
\paragraph{Reinforcement learning with action sequence}
In the context of RL, \citet{medini2019mimicking} propose to pre-compute frequent action sequences from expert demonstrations and augment the action space with these sequences.
However, this idea introduces additional complexity and is not scalable to setups without demonstrations.
One recent work relevant to ours is \citet{saanum2024reinforcement} that encourage a sequence of actions from RL agents to be predictable and smooth.
Our work differs in that we directly incorporate action sequences into value learning.
Recently, \citet{ankile2024imitation} point out that RL with action sequence is challenging and instead propose to use RL for learning a single-step policy that corrects action sequence predictions from BC.
In contrast, we show that RL with action sequence is feasible and improves performance of RL algorithms.

\vspace{-0.075in}
\paragraph{Multi-token prediction}
Recent large language models have incorporated a notably similar idea to predicting action sequences from demonstrations -- predicting multiple future tokens at once, or multi-token prediction \citep{gloeckle2024better,liu2024deepseek}.
For instance, \citet{gloeckle2024better} show that predicting multiple $n$ future tokens in parallel with $n$ independent output heads improves the performance and can accelerate inference speed.
DeepSeek-V3 \citep{liu2024deepseek} also make a similar observation but with a sequential multi-token prediction.
It would be interesting to see whether our idea can be utilized for fine-tuning these models with multi-token prediction.

\vspace{-0.075in}
\paragraph{Hierarchical reinforcement learning} 
Approaches that learn RL agents with temporally extended high-level actions, or options, have been well studied \citep{sutton1988learning}.
The key idea is to train high-level policies that output options by manually defining subgoals \citep{kulkarni2016hierarchical,dayan1992feudal} or learning options from data \citep{bacon2017option,vezhnevets2017feudal,nachum2018data}, and then train a low-level agent that learns to execute low-level actions conditioned on options.
Our work is not directly comparable to these works as we do not abstract temporally extended actions but use raw action sequences for value learning.

\vspace{-0.05in}
\section{Discussion}
\vspace{-0.05in}
\label{sec:discussion}
We have presented Coarse-to-fine Q-Network with \textbf{A}ction \textbf{S}equence (CQN-\textbf{AS}), a critic-only RL algorithm that trains a critic network to output Q-values over action sequences.
Extensive experiments in benchmarks with various setups show that our idea not only improves the performance of the base algorithm but also allows for solving complex tasks where prior RL algorithms fail.

\vspace{-0.075in}
\paragraph{Limitations and future work}
One limitation of our work is the lack of real-world robot evaluation.
Moreover, as discussed in \cref{sec:bigym_experiments} and \cref{sec:ablation_analysis_failure}, solving tasks involving small objects remains a limitation of our approach.
One potential approach would be using strong pre-trained vision encoders, but we find that computational cost is often prohibitively large, which remains as an open problem.
We are excited about future directions, including real-world RL with humanoid robots, incorporating advanced critic architectures \citep{kapturowski2022human,chebotar2023q,springenberg2024offline}, bootstrapping RL agents from imitation learning \citep{hu2023imitation,xing2024bootstrapping} or offline RL \citep{nair2020awac,lee2021offline}, extending the idea to recent model-based RL approaches \citep{hafner2023mastering,hansen2023td}, extend parallel value learning scheme to autoregressive, multi-step Q-learning scheme \citep{kahn2018self}, fine-tuning vision-language-action models that use action sequence \citep{team2024octo,doshi2024scaling} or language models that use multi-token prediction \citep{gloeckle2024better,liu2024deepseek} with our algorithm, to name but a few.

\newpage

\section*{Acknowledgements}
We thank Stephen James and Richie Lo for the discussion on the initial idea of this project.
This work was supported in part by Multidisciplinary University Research Initiative (MURI) award by the Army Research Office (ARO) grant No. W911NF-23-1-0277.
Pieter Abbeel holds concurrent appointments as a Professor at UC Berkeley and as an Amazon Scholar.
This paper describes work performed at UC Berkeley and is not associated with Amazon.

\bibliography{reference}
\bibliographystyle{ref_bst}

\clearpage

\appendix

\section{Motivating Experiments}
\label{appendix:motivating_experiments}

\paragraph{Additional details} 
For \cref{fig:return_to_go_prediction}, we use the same demonstrations used in our main experiments (see \cref{appendix:experimental_details} for more details).
For SAC and TD3 experiments with action sequences in \cref{fig:value_overestimation}, we implemented our code based on the official HumanoidBench repository. We use the hyperparameters in the repository for training SAC agents.
For TD3, we use the standard deviation of 0.2 for exploration.
We report the average target Q-values recorded throughout experiments.
For \cref{fig:no_op_actions}, we train SAC and CQN agents with 6 original actions and 294 no-op actions with [-1, 1] action bounds and use an environment wrapper that slices out no-op actions.

\begin{figure*}[h]
    \centering
    \subfloat[2D Point-mass task]{
    \includegraphics[width=0.48\linewidth]{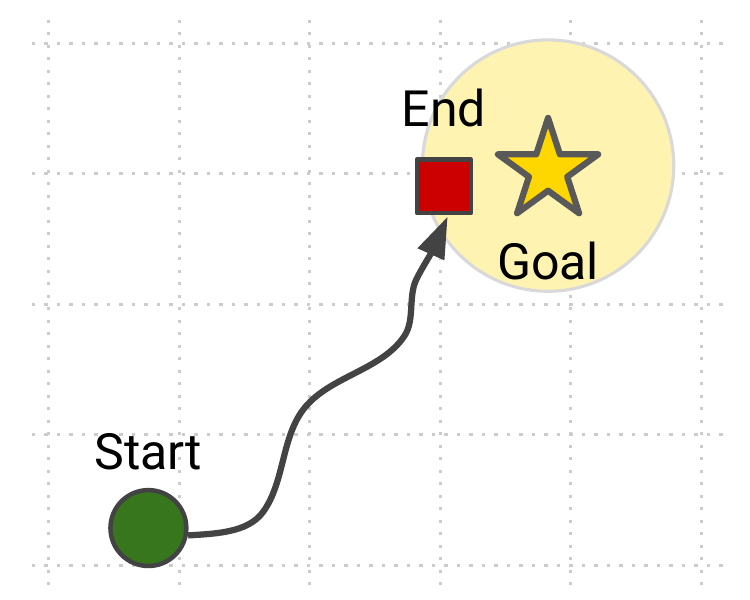}
    \label{fig:point_mass_visualization}
    }
    \subfloat[Experimental Results]{
    \includegraphics[width=0.48\linewidth]{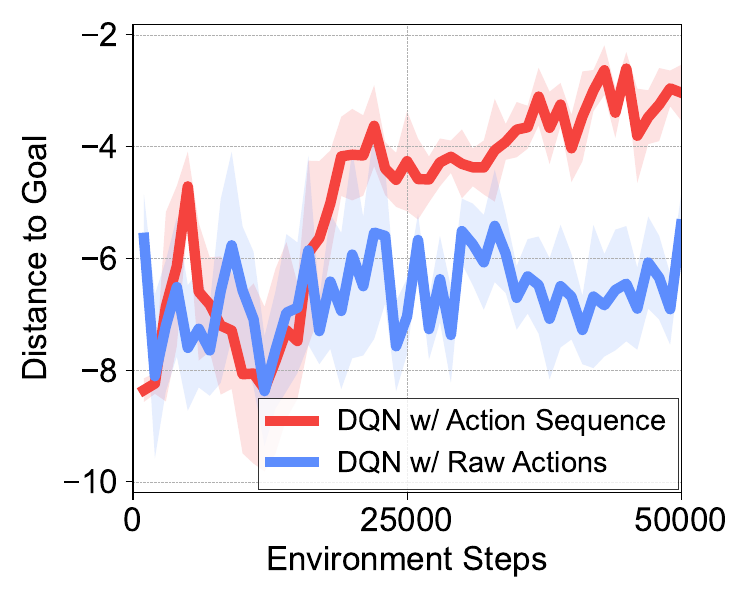}
    \label{fig:point_mass_results}
    }
    \caption{\textbf{2D Point-mass experiments} (a) We consider a simple toy environment where the goal is to reach the region around a randomly spawned goal point. (b) We show that, as expected, training with pre-defined action sequences lead to faster convergence.
    The solid line and shaded regions represent the mean and confidence intervals, respectively, across 10 runs.}
    \label{fig:point_mass}
    \end{figure*}

\paragraph{Experiment with 2D Point-mass environment}
To further motivate the use of action sequence for value learning, we train DQN agents \citep{mnih2015human} on 2D Point-mass environment with discrete action spaces. In particular, we train a \texttt{Raw} agent that trains with the raw discrete action space consisting of single-step accelerations parameterized by 8 discrete headings (cardinal and discrete directions) and 1 magnitude level, resulting in 8 total actions.
We compare this against a \texttt{Sequence} agent that trains with the discrete action space that consists of smooth 5-step acceleration sequences parameterized by cubic Bezier curves instead of single-step accelerations. Both agents operate on a 2D double-integrator environment where the goal is to reach a target position.
In \cref{fig:point_mass}, as expected, we find that training DQN agent with pre-defined action sequences lead to faster convergence.
Our main experimental results in \cref{sec:experiment} further show that, CQN-AS can achieve similar benefit of using action sequences without pre-defined set of action sequences on various challenging continuous control benchmarks.

\newpage

\section{Experimental Details}
\label{appendix:experimental_details}

\vspace{-0.025in}
\paragraph{BiGym}
BiGym\footnote{\url{https://github.com/chernyadev/bigym}} \citep{chernyadev2024bigym} is built upon MuJoCo \citep{todorov2012mujoco}.
We use Unitree H1 with two parallel grippers.
We find that demonstrations available in the recent version of BiGym are not all successful.
Therefore we adopt the strategy of replaying all the demonstrations and only use the successful ones as demonstrations.
instead of discarding the failed demonstrations, we store them in a replay buffer as failure experiences.
To avoid training with too few demonstrations,
we exclude the tasks where the ratio of successful demonstrations is below 50\%.
\cref{table:bigym_tasks} shows the list of 25 sparsely-rewarded mobile bi-manual manipulation tasks used in our experiments.

\begin{table}[ht]
\vspace{-0.025in}
\caption{\textbf{BiGym tasks} with their maximum episode length and number of successful demonstrations.}
\small
\begin{center}
\begin{tabular}{lcclcc}
\toprule
\textbf{Task} & \textbf{Length} & \textbf{Demos} & \textbf{Task} & \textbf{Length} & \textbf{Demos} \\
\midrule
Move Plate & 300 & 51 & Cupboards Close All & 620 & 53 \\
Move Two Plates & 550 & 30 & Reach Target Single & 100 & 30 \\
Saucepan To Hob  & 440 & 28 & Reach Target Multi Modal & 100 & 60 \\
Sandwich Flip & 620 & 34 & Reach Target Dual & 100 & 50 \\
Sandwich Remove & 540 & 24 & Dishwasher Close & 375 & 44 \\
Dishwasher Load Plates & 560 & 17 & Wall Cupboard Open  & 300 & 44 \\
Dishwasher Load Cups & 750 & 58 & Drawers Open All & 480 & 45 \\
Dishwasher Unload Cutlery & 620 & 29 & Wall Cupboard Close & 300 & 60 \\
Take Cups & 420 & 32 & Dishwasher Open Trays & 380 & 57 \\
Put Cups & 425 & 43 & Drawers Close All & 200 & 59 \\
Flip Cup & 550 & 45 & Drawer Top Open & 200 & 40 \\
Flip Cutlery & 500 & 43 & Drawer Top Close & 120 & 51 \\
Dishwasher Close Trays & 320 & 62 & & & \\
\bottomrule
\end{tabular}
\label{table:bigym_tasks}
\end{center}
\end{table} 

\vspace{-0.025in}
\paragraph{HumanoidBench}
HumanoidBench\footnote{\url{https://github.com/carlosferrazza/humanoid-bench}} \citep{sferrazza2024humanoidbench} is built upon MuJoCo \citep{todorov2012mujoco}.
We use Unitree H1 with two dexterous hands.
We consider the first 8 locomotion tasks in the benchmark: \texttt{Stand}, \texttt{Walk}, \texttt{Run}, \texttt{Reach}, \texttt{Hurdle}, \texttt{Crawl}, \texttt{Maze}, \texttt{Sit} \texttt{Simple}.
We use proprioceptive states and privileged task information instead of visual observations.
Unlike BiGym and RLBench experiments, we do not utilize dueling network \citep{wang2016dueling} and distributional critic \citep{bellemare2017distributional} in HumanoidBench for faster experimentation.

\vspace{-0.025in}
\paragraph{RLBench}
RLBench\footnote{\url{https://github.com/stepjam/RLBench}} \citep{james2020rlbench} is built upon CoppeliaSim \citep{rohmer2013v} and PyRep \citep{james2019pyrep}.
We use a 7-DoF Franka Panda robot arm and a parallel gripper.
Following the setup of \citet{seo2024continuous}, we increase the velocity and acceleration of the arm by 2 times.
For all experiments, we use 100 demonstrations generated via motion-planning.
\cref{table:rlbench_tasks} shows the list of 20 sparsely-rewarded visual manipulation tasks used in our experiments.

\begin{table}[ht]
\vspace{-0.025in}
\caption{\textbf{RLBench tasks} with their maximum episode length used in our experiments.}
\small
\begin{center}
\begin{tabular}{lclc}
\toprule
\textbf{Task} & \textbf{Length} & \textbf{Task} & \textbf{Length}  \\
\midrule
Take Lid Off Saucepan & 100 & Put Books On Bookshelf  & 175 \\
Open Drawer  & 100 & Sweep To Dustpan & 100 \\
Stack Wine  & 150 & Pick Up Cup & 100 \\
Toilet Seat Up & 150 & Open Door  & 125 \\
Open Microwave & 125 & Meat On Grill  & 150 \\
Open Oven & 225 & Basketball In Hoop  & 125 \\
\makecell[l]{Take Plate Off\\Colored Dish Rack}  & 150 & Lamp On  & 100 \\
Turn Tap & 125 & Press Switch  & 100 \\
Put Money In Safe & 150 & Put Rubbish In Bin  & 150 \\
Phone on Base & 175 & Insert Usb In Computer  & 100 \\
\bottomrule
\end{tabular}
\label{table:rlbench_tasks}
\end{center}
\end{table} 

\newpage

\paragraph{Hyperparameters}
We use the same set of hyperparameters across the tasks in each domain.
For hyperparameters shared across CQN and CQN-\textbf{AS}, we use the same hyperparameters for both algorithms for a fair comparison.
We provide detailed hyperparameters for BiGym and RLBench experiments in \cref{table:hyperparameters_bigym_rlbench} and HumanoidBench experiments in \cref{table:hyperparameters_humanoidbench}

\begin{table}[ht]
\caption{Hyperparameters for demo-driven vision-based experiments in BiGym and RLBench}
\small
\begin{center}
\begin{tabular}{ll}
\toprule
\textbf{Hyperparameter} & \textbf{Value}  \\
\midrule
Image resolution  & $84 \times 84 \times 3$ \\
Image augmentation & RandomShift \citep{yarats2022mastering} \\
Frame stack & 4 (BiGym) / 8 (RLBench) \\\midrule
CNN - Architecture & Conv (c=[32, 64, 128, 256], s=2, p=1) \\
MLP - Architecture & \makecell[l]{Linear (c=[512, 512, 64, 512, 512], bias=False) (BiGym) \\Linear (c=[64, 512, 512], bias=False) (RLBench)} \\
CNN \& MLP - Activation & SiLU \citep{hendrycks2016gaussian} and LayerNorm \citep{ba2016layer} \\
GRU - Architecture & GRU (c=[512], bidirectional=False) \\
Dueling network & True  \\\midrule
C51 - Atoms & 51 \\
C51 - $\text{v}_{\text{min}}$, $\text{v}_{\text{max}}$ & -2, 2 \\\midrule
Action sequence & 16 (BiGym) / 4 (RLBench) \\
Temporal ensemble weight $m$ & 0.01 \\
Levels & 3 \\
Bins & 5 \\ \midrule
BC loss ($\mathcal{L}_{\tt{BC}}$) scale & 1.0 \\
RL loss ($\mathcal{L}_{\tt{RL}}$) scale & 0.1 \\
Relabeling as demonstrations & True \\
Data-driven action scaling & True \\
Action mode & Absolute Joint (BiGym), Delta Joint (RLBench) \\ 
Exploration noise & $\epsilon \sim \mathcal{N}(0, 0.01)$ \\
Target critic update ratio ($\tau$) & 0.02 \\
N-step return & 1 \\
Batch size & 256 \\
Demo batch size & 256 \\
Optimizer & AdamW \citep{loshchilov2018decoupled} \\
Learning rate & 5e-5 \\
Weight decay & 0.1 \\
\bottomrule
\end{tabular}
\label{table:hyperparameters_bigym_rlbench}
\end{center}
\vspace{-0.1in}
\end{table} 
\begin{table}[ht]
\caption{Hyperparameters for state-based experiments in HumanoidBench}
\small
\begin{center}
\begin{tabular}{ll}
\toprule
\textbf{Hyperparameter} & \textbf{Value}  \\
\midrule 
MLP - Architecture & Linear (c=[512, 512], bias=False) \\
CNN \& MLP - Activation & SiLU \citep{hendrycks2016gaussian} and LayerNorm \citep{ba2016layer} \\
GRU - Architecture & GRU (c=[512], bidirectional=False) \\
Dueling network & True \\\midrule
Action sequence & 4 \\
Temporal ensemble weight $m$ & 0.01 \\
Levels & 3 \\
Bins & 5 \\ \midrule
RL loss ($\mathcal{L}_{\tt{RL}}$) scale & 1.0 \\
Action mode & Absolute Joint \\ 
Exploration noise & $\epsilon \sim \mathcal{N}(0, 0.01)$ \\
Target critic update ratio ($\tau$) & 1.0 \\
Target critic update interval ($\tau$) & 100 \\
Update-to-data ratio (UTD) & 0.5 \\
N-step return & 3 \\
Batch size & 128 \\
Optimizer & AdamW \citep{loshchilov2018decoupled} \\
Learning rate & 5e-5 \\
Weight decay & 0.1 \\
\bottomrule
\end{tabular}
\label{table:hyperparameters_humanoidbench}
\end{center}
\end{table} 

\paragraph{Computing hardware}
For BiGym and Humanoid experiments, we use NVIDIA A5000 GPU with 24GB VRAM. With A5000, each BiGym experiment with 100K environment steps take 16 hours, and each HumanoidBench experiment with 10M environment steps take 40 hours. For RLBench experiments, we use NVIDIA RTX 2080Ti GPU, with which each experiment with 30K environment steps take 6.5 hours.
We find that CQN-AS takes 13\% more memory compared to CQN and is 40\% slower than CQN.
Overall, CQN-\textbf{AS} is around 33\% slower than running CQN because larger architecture slows down both training and inference.

\paragraph{Baseline implementation}
For CQN \citep{seo2024continuous} and DrQ-v2+ \citep{yarats2022mastering}, we use the implementation available from the official CQN implementation\footnote{\url{https://github.com/younggyoseo/CQN}}.
For ACT \citep{zhao2023learning}, we use the implementation from RoboBase repository\footnote{\url{https://github.com/robobase-org/robobase}}.
For SAC \citep{haarnoja2018soft}, DreamerV3 \citep{hafner2023mastering}, and TD-MPC2 \citep{hansen2023td}, we use results provided in HumanoidBench\footnote{\url{https://github.com/carlosferrazza/humanoid-bench}} repository \citep{sferrazza2024humanoidbench}.

\newpage

\section{Full description of CQN and CQN-AS}
\label{appendix:full_description_cqn_as}

This section provides the formulation of CQN and CQN-\textbf{AS} with $n$-dimensional actions.

\subsection{Coarse-to-fine Q-Network}
Let $a_{t}^{l,n}$ be an action at level $l$ and dimension $n$ and $\mathbf{a}_{t}^{l} = \{a_{t}^{l,1}, ..., a_{t}^{l,N}\}$ be actions at level $l$ with $\mathbf{a}_{t}^{0}$ being zero vector.
We then define coarse-to-fine critic to consist of multiple Q-networks:
\begin{align}
    Q_{\theta}^{l, n}(\mathbf{h}_{t}, a_{t}^{l,n}, \mathbf{a}_{t}^{l-1}) = \left[Q^l_\theta(\mathbf{h}_t, a^{l,n}_t = a^{l,n,b}_t, a^{l-1}_t) \right]_{b=1}^{B} \; \text{for} \; l \in \{1, ..., L\} \; \text{and} \; n \in \{1, ..., N\}
\end{align}
Where $B$ denotes the number of bins. We optimize the critic network with the following objective:
\begin{align}
    \sum\nolimits_{n}\sum\nolimits_{l}\left( Q_{\theta}^{l,n}(\mathbf{h}_{t}, a_{t}^{l,n}, \mathbf{a}_{t}^{l-1}) - r_{t+1} - \gamma \max_{a'}Q_{\bar{\theta}}^{l,n}(\mathbf{h}_{t+1}, a', \pi^{l-1}(\mathbf{h}_{t+1}) \right)^{2},
\end{align}
where $\bar{\theta}$ are delayed parameters for a target network \citep{polyak1992acceleration} and $\pi^{l}$ is a policy that outputs the action $\mathbf{a}_{t}^{l}$ at each level $l$ via the inference steps with our critic, \textit{i.e.,} $\pi^{l}(\mathbf{h}_{t}) = \mathbf{a}_{t}^{l}$.

\paragraph{Action inference}
To output actions at time step $t$ with the critic, CQN first initializes constants $a_{t}^{n, \tt{low}}$ and $a_{t}^{n, \tt{high}}$ with $-1$ and $1$ for each $n$.
Then the following steps are repeated for $l \in \{1, ..., L\}$:
\begin{itemize}
    \item Step 1 (Discretization): Discretize an interval $[a_{t}^{n, \tt{low}}, a_{t}^{n, \tt{high}}]$ into $B$ uniform intervals, and each of these intervals become an action space for $Q_{\theta}^{l,n}$.
    \item Step 2 (Bin selection): Find the bin with the highest Q-value, set $a_{t}^{l,n}$ to the centroid of the selected bin, and aggregate actions from all dimensions to $\mathbf{a}_{t}^{l}$.
    \item Step 3 (Zoom-in): Set $a_{t}^{n, \tt{low}}$ and $a_{t}^{n, \tt{high}}$ to the minimum and maximum of the selected bin, which intuitively can be seen as zooming-into each bin.
\end{itemize}
We then use the last level's action $\mathbf{a}_{t}^{L}$ as the action at time step $t$.

\newpage

\paragraph{Computing Q-values}
To compute Q-values for given actions $\mathbf{a}_{t}$, CQN first initializes constants $a_{t}^{n, \tt{low}}$ and $a_{t}^{n, \tt{high}}$ with $-1$ and $1$ for each $n$.
We then repeat the following steps for $l \in \{1, ..., L\}$:
\begin{itemize}
    \item Step 1 (Discretization): Discretize an interval $[a_{t}^{n, \tt{low}}, a_{t}^{n, \tt{high}}]$ into $B$ uniform intervals, and each of these intervals become an action space for $Q_{\theta}^{l,n}$.
    \item Step 2 (Bin selection): Find the bin that contains input action $\mathbf{a}_{t}$, compute $a^{l,n}_{t}$ for the selected interval, and compute Q-values $Q_{\theta}^{l,n}(\mathbf{h}_{t}, a^{l,n}_{t}, \mathbf{a}_{t}^{l-1})$.
    \item Step 3 (Zoom-in): Set $a_{t}^{n, \tt{low}}$ and $a_{t}^{n, \tt{high}}$ to the minimum and maximum of the selected bin, which intuitively can be seen as zooming-into each bin.
\end{itemize}
We then use a set of Q-values $\{Q_{\theta}^{l,n}(\mathbf{h}_{t}, a^{l,n}_{t}, \mathbf{a}_{t}^{l-1})\}_{l=1}^{L}$ for given actions $\mathbf{a}_{t}$.

\subsection{Coarse-to-fine Critic with Action Sequence}

Let $\mathbf{a}_{t:t+K}^{l} = \{ \mathbf{a}_{t}^{l}, ..., \mathbf{a}_{t+K-1}^{l}\}$ be an action sequence at level $l$ and $\mathbf{a}_{t:t+K}^{0}$ be zero vector.
Our critic network consists of multiple Q-networks for each level $l$, dimension $n$, and sequence step $k$:
\begin{gather}
\begin{aligned}
    Q_{\theta}^{l, n, k}(\mathbf{h}_{t}, a_{t+k-1}^{l,n}, \mathbf{a}_{t:t+K}^{l-1}) = \left[Q_{\theta}^{l, n, k}(\mathbf{h}_{t}, a_{t+k-1}^{l,n}=a_{t+k-1}^{l,n,b}, \mathbf{a}_{t:t+K}^{l-1})\right]_{b=1}^{B} \\ \text{for} \;\; l \in \{1, ..., L\}, \; n \in \{1, ..., N\} \; \text{and} \; k \in \{1, ..., K\}
\end{aligned}
\end{gather}
We optimize the critic network with the following objective:
\begin{align}
    \sum_{n}\sum_{l}\sum_{k} \left( Q_{\theta}^{l,n,k}(\mathbf{h}_{t}, a_{t}^{l,n}, \mathbf{a}_{t:t+K}^{l-1}) - r_{t+1} - \gamma \max_{a'}Q_{\bar{\theta}}^{l,n,k}(\mathbf{h}_{t+1}, a', \pi^{l-1}_{K}(\mathbf{h}_{t+1}) \right)^{2},
\end{align}
where $\pi_{K}^{l}$ is an action sequence policy that outputs the action sequence $\mathbf{a}_{t:t+K}^{l}$.
In practice, we compute Q-values for all sequence step $k \in \{1, ..., K\}$ and all action dimension $n \in \{1, ..., N\}$ in parallel.
This can be seen as extending the idea of \citet{seyde2023solving}, which learns decentralized Q-networks for action dimensions, into action sequence dimension.
As we mentioned in \cref{sec:c2f_critic_as}, we find this simple scheme works well on challenging tasks with high-dimensional action spaces.

\paragraph{Architecture}
Let $\mathbf{e}_{k}$ denote an one-hot encoding for $k$.
For each level $l$, we construct features for each sequence step $k$ as $\mathbf{h}^{l}_{t,k} = \left[\mathbf{h}_{t}, \mathbf{a}_{t+k-1}^{l-1}, \mathbf{e}_{k}\right]$.
We encode each $\mathbf{h}_{t,k}^{l}$ with a shared MLP network and process them through GRU \citep{cho2014learning} to obtain $\mathbf{s}^{l}_{t,k} = f^{\tt{GRU}}_{\theta}(f^{\tt{MLP}}_{\theta}(\mathbf{h}_{t,1}^{l}), ..., f^{\tt{MLP}}_{\theta}(\mathbf{h}_{t,k}^{l}))$.
We use a shared projection layer to map each $\mathbf{s}_{t,k}^{l}$ into Q-values at each sequence step $k$, \textit{i.e.,} $f_{\theta}^{\tt{proj}}(\mathbf{s}_{t,k}^{l}) = \{[Q_{\theta}^{l,k}(\mathbf{h}_{t}, a_{t+k-1}^{l,n,b}, \mathbf{a}_{t:t+K}^{l-1})]_{b=1}^{B}\}_{n=1}^{N}$.
We compute Q-values for all dimensions $n \in \{1, ..., N\}$ at the same time with a big linear layer, which follows the design of \citet{seo2024continuous}.

\section{Additional Preliminary Experiments}

\paragraph{Offline RL with CQN-AS}
To further investigate whether CQN-AS formulation is compatible with offline RL, we conduct preliminary experiments on BiGym's \texttt{Sandwich Remove} task. 
Specifically, we combine CQN-AS with Cal-QL \citep{nakamoto2023cal} and train it on the dataset that consists of 26 successful demonstrations and 10 failed trajectories.
We find that CQN-AS + Cal-QL achieves 33 ($\pm$ 6.8) \% while CQN + Cal-QL achieves 7 ($\pm$ 14) \%, which shows CQN-AS can be indeed effective for offline setup.
We leave further exhaustive investigation as an interesting future work.

\paragraph{Experiments with ResNet-18}
To investigate if using larger and stronger pre-trained vision encoders such as ResNet \citep{he2016deep} can improve performance, we tried running CQN-AS with ResNet-18 encoder on BiGym tasks. However, we find that it requires a GPU with at least 48GB memory and is extremely slow to train. We will leave this direction of incorporating larger vision encoders in an efficient manner as a future direction.

\newpage
\section*{NeurIPS Paper Checklist}

\begin{enumerate}

\item {\bf Claims}
    \item[] Question: Do the main claims made in the abstract and introduction accurately reflect the paper's contributions and scope?
    \item[] Answer: \answerYes{}
    \item[] Justification: We have claimed that incorporating action sequences into reinforcement learning can be beneficial in the abstract and introduction, which clearly reflects the main contribution of this paper -- CQN-\textbf{AS} that uses action sequences in value learning. We have provided experimental results and analysis in the paper to support the claims in \cref{fig:analysis}, \cref{fig:bigym_experiments}, \cref{fig:rlbench_experiments}, and \cref{fig:ablation_and_analysis}.

\item {\bf Limitations}
    \item[] Question: Does the paper discuss the limitations of the work performed by the authors?
    \item[] Answer: \answerYes{}
    \item[] Justification: We have discussed our limitations in \cref{sec:bigym_experiments}, \cref{sec:ablation_analysis_failure}, and \cref{sec:discussion}.

\item {\bf Theory assumptions and proofs}
    \item[] Question: For each theoretical result, does the paper provide the full set of assumptions and a complete (and correct) proof?
    \item[] Answer: \answerNA{}
    \item[] Justification: \answerNA{}

    \item {\bf Experimental result reproducibility}
    \item[] Question: Does the paper fully disclose all the information needed to reproduce the main experimental results of the paper to the extent that it affects the main claims and/or conclusions of the paper (regardless of whether the code and data are provided or not)?
    \item[] Answer: \answerYes{}
    \item[] Justification: We have provided the detailed experimental details along with the source code to reproduce the results.

\item {\bf Open access to data and code}
    \item[] Question: Does the paper provide open access to the data and code, with sufficient instructions to faithfully reproduce the main experimental results, as described in supplemental material?
    \item[] Answer: \answerYes{}
    \item[] Justification: We have provided the source code.

\item {\bf Experimental setting/details}
    \item[] Question: Does the paper specify all the training and test details (e.g., data splits, hyperparameters, how they were chosen, type of optimizer, etc.) necessary to understand the results?
    \item[] Answer: \answerYes{}
    \item[] Justification: We have provided sufficient details used in our experiments along with the source code that contains hyperparameters required for reproducing the results.

\item {\bf Experiment statistical significance}
    \item[] Question: Does the paper report error bars suitably and correctly defined or other appropriate information about the statistical significance of the experiments?
    \item[] Answer: \answerYes{}
    \item[] Justification: We have provided error bars aggregated over 8 runs for our experiments.

\item {\bf Experiments compute resources}
    \item[] Question: For each experiment, does the paper provide sufficient information on the computer resources (type of compute workers, memory, time of execution) needed to reproduce the experiments?
    \item[] Answer: \answerYes{}
    \item[] Justification: We have provided detailed information about compute costs in Appendix.
    
\item {\bf Code of ethics}
    \item[] Question: Does the research conducted in the paper conform, in every respect, with the NeurIPS Code of Ethics \url{https://neurips.cc/public/EthicsGuidelines}?
    \item[] Answer: \answerYes{}
    \item[] Justification: We are committed to conforming with the NeurIPS Code of Ethics.

\item {\bf Broader impacts}
    \item[] Question: Does the paper discuss both potential positive societal impacts and negative societal impacts of the work performed?
    \item[] Answer: \answerYes{}
    \item[] Justification: We have discussed broader impacts of this work in \cref{appendix:broader_impacts}.
    
\item {\bf Safeguards}
    \item[] Question: Does the paper describe safeguards that have been put in place for responsible release of data or models that have a high risk for misuse (e.g., pretrained language models, image generators, or scraped datasets)?
    \item[] Answer: \answerNA{}
    \item[] Justification: Our work do not require safeguards as we do not foresee the imminent risks from using our algorithms.

\item {\bf Licenses for existing assets}
    \item[] Question: Are the creators or original owners of assets (e.g., code, data, models), used in the paper, properly credited and are the license and terms of use explicitly mentioned and properly respected?
    \item[] Answer: \answerYes{}
    \item[] Justification: We have cited references for benchmarks we used in the paper. Each benchmark provides licenses for assets used in their simulations in their papers or websites.

\item {\bf New assets}
    \item[] Question: Are new assets introduced in the paper well documented and is the documentation provided alongside the assets?
    \item[] Answer: \answerNA{}
    \item[] Justification: We have not introduced new assets.

\item {\bf Crowdsourcing and research with human subjects}
    \item[] Question: For crowdsourcing experiments and research with human subjects, does the paper include the full text of instructions given to participants and screenshots, if applicable, as well as details about compensation (if any)? 
    \item[] Answer: \answerNA{}
    \item[] Justification: We have not conducted relevant experiments in this paper.

\item {\bf Institutional review board (IRB) approvals or equivalent for research with human subjects}
    \item[] Question: Does the paper describe potential risks incurred by study participants, whether such risks were disclosed to the subjects, and whether Institutional Review Board (IRB) approvals (or an equivalent approval/review based on the requirements of your country or institution) were obtained?
    \item[] Answer: \answerNA{}
    \item[] Justification: We have not conducted relevant experiments in this work.

\item {\bf Declaration of LLM usage}
    \item[] Question: Does the paper describe the usage of LLMs if it is an important, original, or non-standard component of the core methods in this research? Note that if the LLM is used only for writing, editing, or formatting purposes and does not impact the core methodology, scientific rigorousness, or originality of the research, declaration is not required.
    \item[] Answer: \answerNA{}
    \item[] Justification: Developing our algorithm have not involved LLMs.

\end{enumerate}

\end{document}